\documentclass[12pt]{article}
\usepackage{amsmath}
\usepackage{graphicx,psfrag,epsf}
\usepackage{enumerate}
\usepackage{natbib}
\usepackage{url} 
\usepackage{algorithm}
\usepackage{algpseudocode}
\usepackage{tabularray}
\usepackage{multirow}
\usepackage{comment}
\usepackage{enumitem}
\usepackage{authblk}
\usepackage{caption}
\usepackage{tikz}
\usepackage{threeparttable}

\usepackage{hyperref}       
\usepackage{url}            
\usepackage{booktabs}       
\usepackage{amsfonts}       
\usepackage{mathrsfs}
\usepackage{nicefrac}       
\usepackage{microtype}      
\usepackage{xcolor}         
\usepackage{graphicx}
\usepackage{epstopdf}
\usepackage{float}
\usepackage{subfigure}
\usepackage{amssymb}
\usepackage{adjustbox}
\usepackage{bbding}
\usepackage{amsmath}
\usepackage{setspace}
\usepackage{color}
\usepackage{pifont}
\newcommand{\blind}{0}

\begin{document}

\def\spacingset#1{\renewcommand{\baselinestretch}%
{#1}\small\normalsize} \spacingset{1}


\if0\blind
{
  \title{\LARGE\bf LAMBDA: A Large Model Based Data Agent}
	\author{\small Maojun $\rm{Sun}^{b}$,  Ruijian $\rm{Han}^{b}$,
	Binyan $\rm{Jiang}^{b}$, \\
	Houduo  $\rm{Qi}^{a,b}$,
	Defeng $\rm{Sun}^{a}$,
	Yancheng $\rm{Yuan}^{a*}$
	and
	Jian $\rm{Huang}^{a,b}$\thanks{Corresponding authors.}
	
	\vspace{0.2cm}
	{\footnotesize {{$\it^{a}$Department of Applied Mathematics, The Hong Kong Polytechnic University}}}\\
	{\footnotesize { {$\it^{b}$Department of Data Science and Artificial Intelligence, The Hong Kong Polytechnic University}}}
}
\date{\footnotesize \today}
  \maketitle} \fi

\if1\blind
{
  \bigskip
  \bigskip
  \bigskip
  \begin{center}
    {\LARGE\bf LAMBDA: A Large Model Based Data Agent}
\end{center}
  \medskip
} \fi

\bigskip
\begin{abstract}
We introduce LArge Model Based Data Agent (LAMBDA), a novel open-source, code-free multi-agent data analysis system that leverages the power of large language models. LAMBDA is designed to address data analysis challenges in data-driven applications through innovatively designed data agents using natural language. At the core of LAMBDA are two key agent roles: the programmer and the inspector, which are engineered to work together seamlessly. Specifically, the programmer generates code based on the user's instructions and domain-specific knowledge, while the inspector debugs the code when necessary. To ensure robustness and handle adverse scenarios, LAMBDA features a user interface that allows direct user intervention. Moreover, LAMBDA can flexibly integrate external models and algorithms through our proposed Knowledge Integration Mechanism, catering to the needs of customized data analysis. LAMBDA has demonstrated strong performance on various data analysis tasks. It has the potential to enhance data analysis paradigms by seamlessly integrating human and artificial intelligence, making it more accessible, effective, and efficient for users from diverse backgrounds. The strong performance of LAMBDA in solving data analysis problems is demonstrated using real-world data examples.
The code for LAMBDA is available at \url{https://github.com/AMA-CMFAI/LAMBDA}
and videos of three case studies can be viewed at \url{https://www.polyu.edu.hk/ama/cmfai/lambda.html}.
\end{abstract}

\noindent%
\vfill

\newpage
\spacingset{1.5} 
\section{Introduction}
Over the past decade, the data-driven approach utilizing deep neural networks has driven the success of artificial intelligence across many challenging applications in various fields \citep{lecun2015deep}. Despite these advancements, the current paradigm encounters challenges and limitations in statistical and data science applications, particularly in domains such as biology \citep{weissgerber2016reinventing}, healthcare \citep{oakes2024building}, and business \citep{weihs2018data}, which require extensive expertise and advanced coding knowledge for data analysis. A significant barrier is the lack of effective communication channels between domain experts and sophisticated AI models \citep{park2021facilitating}. To address this issue, we introduce a Large Model Based Data Agent (LAMBDA), which is a new open-source, code-free multi-agent data analysis system designed to overcome this dilemma. LAMBDA aims to create a much-needed medium, fostering seamless interaction between domain knowledge and the capabilities of AI in statistics and data science.

Our main objectives in developing LAMBDA are as follows.

\textbf{(a) Crossing coding barrier}:
Coding has long been recognized as a significant barrier for domain experts without a background in statistics or computer science, preventing them from effectively leveraging powerful AI tools for data analysis \citep{oakes2024building}. LAMBDA addresses this challenge by enabling users to interact with data agents through natural language instructions, thereby offering a coding-free experience. This approach significantly lowers the barriers to entry for tasks in data science, such as data analysis and data mining, while simultaneously enhancing efficiency and making these tasks more accessible to professionals across various disciplines.

\textbf{(b) Integrating human intelligence and AI}:
The existing paradigm of data analysis is confronted with a challenge due to the lack of an efficient intermediary that connects human intelligence with artificial intelligence \citep{park2021facilitating}.
On one hand, AI models often lack an understanding of the unlearned
domain knowledge required for specific tasks. On the other hand, domain experts find it challenging to integrate their expertise into AI models to enhance their performance \citep{dash2022review}. LAMBDA provides a possible solution to alleviate this problem. With a well-designed interface in our key-value (KV) knowledge base, the agents can access external resources like algorithms or models. This integration ensures that domain-specific knowledge is effectively incorporated, meets the need for customized data analysis, and enhances the agent's ability to perform complex tasks with higher accuracy and relevance.

\textbf{(c) Reshaping data science education}:
LAMBDA has the potential to become an interactive platform that can transform statistical and data science education. It offers educators the flexibility to tailor their teaching plans and seamlessly integrate the latest research findings. This adaptability makes LAMBDA an invaluable tool for educators seeking to provide cutting-edge, personalized learning experiences.
Such an approach stands in contrast to the direct application of models like GPT-4 \citep{openai2024gpt4,Tu2024What}, offering a unique and innovative educational platform.

Beyond these features, the design of LAMBDA also emphasizes reliability and portability. Reliability refers to LAMBDA's ability to handle data analysis tasks stably and automatically address failures. Portability ensures that LAMBDA is compatible with various large language models (LLMs), allowing it to be continuously enhanced by the latest state-of-the-art models. To save users time on tasks such as document writing, LAMBDA is equipped with the capability for automatic analysis report generation. To accommodate diverse user needs, LAMBDA also supports exporting code to IPython notebook files, such as ``ipynb" files in Jupyter Notebook.

While GPT-4 has demonstrated state-of-the-art performance in advanced data analysis, its closed-source nature constrains its adaptability to the rapidly expanding needs of statistical and data science applications and specialized educational fields. Furthermore, concerns regarding data privacy and security risks are inherent in the present configuration of GPT-4 \citep{bavli2024ethical}.
In contrast, by utilizing the open-source LAMBDA, users can alleviate concerns about data privacy by preventing the transmission of user data to external servers. Additionally, it offers greater flexibility and convenience in integrating domain knowledge, installing packages, and utilizing various computational resources.

LAMBDA demonstrates exceptional performance across various datasets used in our system testing. Moreover, it outperforms other data agents in handling complex domain tasks during our experiments. In summary, our main contributions are as follows: We propose a well-engineered architecture for an LLM-based data agent that enables natural language-driven data analysis in a conversational manner. Unlike typical end-to-end data agents, our design allows human intervention throughout the process, ensuring adaptability when the agent fails to complete a task or misinterprets user intent. Moreover, we introduce a Knowledge Integration mechanism to effectively handle tasks requiring domain-specific knowledge, providing greater flexibility when misalignment occurs in the knowledge. Its ongoing development has the potential to enhance statistics and data science, making advanced tools more accessible to diverse users.

This paper begins with the background and related works in Section \ref{sec:back}. Section \ref{sec:meth} provides a detailed description of the proposed LAMBDA method. To evaluate its effectiveness, we present our experiments and results in Section \ref{sec:exp_result}. Section \ref{sec:example} demonstrates examples and cases of LAMBDA's application in various scenarios, including data analysis, integration of human intelligence, and interactive education. The paper concludes with a summary in Section \ref{sec:con}. More information and details, including implementation, some discussions, datasets, case studies, and experimental settings, are provided in the Supplementary Materials.

\section{Background and related works}\label{sec:back}
In recent years, the rapid progress in LLMs like GPT-3, GPT-4, PaLM, LLaMA, and Qwen \citep{brown2020language, openai2024gpt4, chowdhery2022palm, touvron2023llama, bai2023qwen} has brought boundless possibilities to the field of artificial intelligence and its applications in many fields, including statistics and data science.
Benefiting from this revolution, LLM-powered agents (LLM agents) are developed to automatically solve problems in various domains like the search engine, software engineering, gaming, and data science \citep{guo2024large, hong2023metagpt, wu2023autogen, zhou2023agents, hong2023metagpt}.

\subsection{LLMs as data analysis agents}
LLM-based data science agent, or data agent, is dedicated to harnessing the power of LLMs to automate data science and analysis tasks \citep{sun2024survey}. For example, GPT-4-Advanced Data Analysis and ChatGLM-Data Analysis can analyze user's data files, perform computations, and generate visualizations \citep{openai2024gpt4}. Some works integrate LLMs into Jupyter Notebooks. For instance, MLCopilot \citep{zhang2023mlcopilot} and Chapter \citep{chapyter}, enable users to interact directly with the notebook, greatly enhancing flexibility. However, they cannot automatically fix errors when they occur and require additional magic commands to support natural language input.

Meanwhile, some researchers focus on designing end-to-end data agents to automate the entire pipeline, including data preprocessing and model evaluation, without human intervention. For example, Data Interpreter \citep{hong2024data} and TaskWeaver \citep{taskweaver} accomplish their tasks through planning and iterative steps. However, current state-of-the-art LLM/VLM-based agents do not reliably automate complete data science workflows \citep{NEURIPS2024_c2f71567}. While fully relying on LLMs for each step reduces human effort, it also significantly increases instability and uncertainty. In addition, if any intermediate step does not align with the user's intent, the process must be repeated, potentially leading to token waste. In contrast, LAMBDA is designed to support a human-agent collaboration mode, allowing for human intervention at any stage of the process if necessary.

Furthermore, these works have not adequately addressed the high degree of user flexibility needed in data analysis, such as the integration of custom algorithms or statistical models according to user preferences. This flexibility is crucial for enhancing data analysis tasks in domain-specific applications and in statistical and data science education. To address this gap, we have designed a Knowledge Integration Mechanism that allows for the easy incorporation of user resources into our agent system.

\subsection{Multi-agent collaboration}
A multi-agent system consists of numerous autonomous agents that collaboratively engage in planning, discussions, and decision-making, mirroring the cooperative nature of human group work in problem-solving tasks \citep{guo2024large}. Each agent has unique capabilities, objectives, and perceptions, operating either independently or collectively to tackle complex tasks or resolve problems \citep{huang2023agentcoder}. { \cite{hong2023metagpt} proposed MetaGPT, modeled after a software company, consisting of agents such as Product Manager, Architect, Project Manager, Engineer, and QA Engineer, efficiently breaking down complex tasks into subtasks involving many agents working together. However, even for simple tasks like data visualization, MetaGPT consume a large number of tokens and require more time. In addition, they generate engineering files that need manual execution and lack the immediacy and interactivity essential for intuitive data analysis. In contrast, LAMBDA simplifies the collaboration process by involving only two agents to simulates data analysis workflows, programmer and inspector respectively, reducing token and time consumption. Moreover, its well-designed user interface allows users to intuitively view the analysis results directly on the screen. A comparison and discussion can be found in the supplement materials.}

\subsection{Knowledge integration}
Addressing tasks that require domain-specific knowledge presents a significant challenge for AI agents \citep{zhang2024raft}. Incorporating knowledge into LLMs through in-context learning (ICL) is a promising strategy for acquiring new information. A well-known technique in this regard is retrieval-augmented generation (RAG) \citep{gao2024retrievalaugmented}, which enhances the accuracy and reduces hallucinations of LLM answers by retrieving external sources \citep{lewis2020retrieval,huang2023survey,borgeaud2022improving,mialon2023augmented}. In RAG, resources are divided into sub-fragments, embedded into vectors, and stored in a vector database. The model first queries this database, identifying document fragments relevant to the user's query based on the similarity. These fragments are then utilized to refine the answers generated by the LLMs through ICL \citep{lewis2020retrieval}. However, deploying a general RAG approach in data analysis introduces specific challenges. First, the user's instructions may not align closely with the relevant code fragments in the representation space, resulting in inaccurate searches. Second, when dealing with extensive code, the agents might struggle to contextualize the correct code segments, where accuracy and completeness are essential for codes and final results.

In addition, custom APIs \citep{hong2024data} can be implemented to handle domain-specific tasks \citep{taskweaver, hong2024data}. For example, systems like Data Interpreter and TaskWeaver invoke the corresponding Tools/Plugins directly within the generated code. Compared to direct parameter-passing, this approach offers greater flexibility in tool usage. However, since the agent cannot access the implementation details of these plugins, it is limited to simple plugin usage and may struggle to resolve misalignment between tools and human instructions when plugin usage is inappropriate.

To address these challenges, we develop a specially designed KV knowledge base with integration methods. This allows users to choose between different modes, including `Full' and `Core', based on the complexity, length of the knowledge context, and specific task requirements. By integrating knowledge, our agent system becomes more adaptable to domain-specific tasks, leveraging human expertise more effectively.

\section{Methodology}
\label{sec:meth}
\begin{figure}[H]
	\begin{center}
		\includegraphics[width=\textwidth]{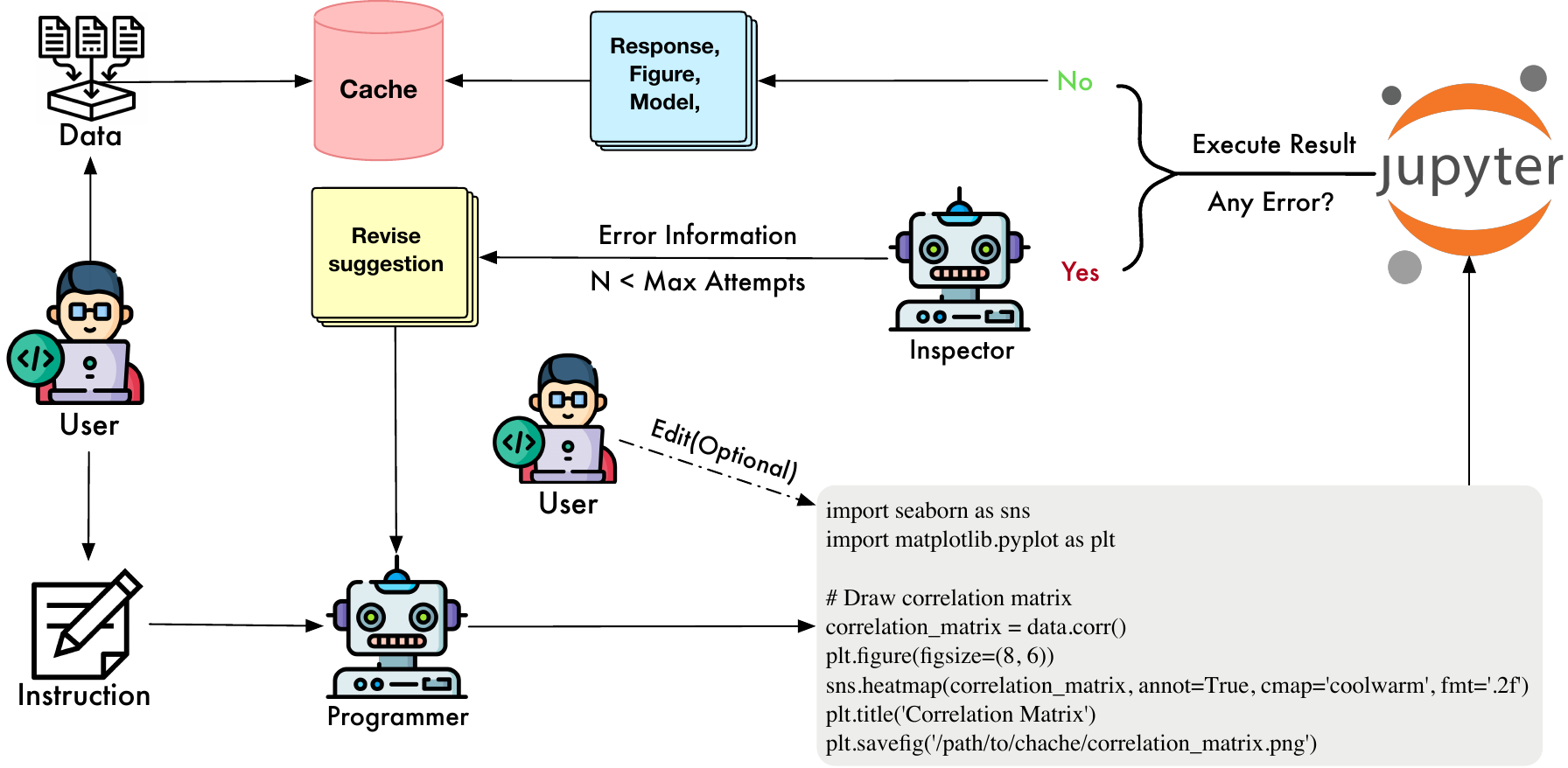}
	\end{center}
\caption{Overview of  LAMBDA. LAMBDA features two core agents: the ``programmer" for code generation and the ``inspector" for error evaluation. The programmer writes and executes code based on user instructions, while the inspector suggests refinements if errors occur. This iterative process continues until the code is error-free or a maximum number of attempts is reached. A human intervention mechanism allows users to modify and run the code directly when needed.
\label{fig:LAMBDA}}
\end{figure}

Our proposed multi-agent data analysis system, LAMBDA, consists of two agents that cooperate seamlessly to solve data analysis tasks using natural language, as shown in Figure \ref{fig:LAMBDA}. The macro workflow describes the code generation process based on user instructions and subsequently executing that code.

\subsection{Overview}
LAMBDA is structured around two core agent roles: the ``programmer" and the ``inspector," who are tasked with code generation and error evaluation, respectively.
The two agents can be implemented separately using either the same or different LLMs.
When users submit an instruction, the programmer agent writes code based on the provided instruction and dataset. This code is then executed within the kernel of the host system. Should any errors arise during execution, the inspector intervenes, offering suggestions for code refinement. The programmer takes these suggestions into account, revises the code, and resubmits it for re-evaluation. This iterative cycle continues until the code runs error-free or a preset maximum number of attempts is reached. In order to cope with adverse
 situations and enhance its reliability and flexibility, a human intervention mechanism is integrated into the workflow. This feature allows users to modify and run the code directly and intervene when necessary. The multi-agent collaboration algorithm is demonstrated in Algorithm \ref{alg:multi_agent}.

\begin{algorithm}
	\caption{Multi-agent Collaboration. 
$A_n$, $C_n$ are the answer and extracted code by the programmer agent in iteration $n$. We assume each $A_n$ contains $C_n$, otherwise, the programmer's reply will be returned to the user directly. $r$ is the execution result,  $E$ indicates an error, $S_n$ are suggestions provided by the inspector in iteration $n$, $C_h$ is the code written by a human. The final response is denoted as $R$.}
	\label{alg:multi_agent}
	\begin{algorithmic}[1]
		\Require $Pr$: Programmer agent
		\Require $I$: Inspector agent
		\Require $d$: Dataset provided by  user
		\Require $ins$: Instructions provided by  user
		\Require $T$: Maximum number of attempts
		\State $n \gets 0$ \Comment{Initialize iteration counter}
		\State $C_n \gets A_n$, $A_n \gets Pr(d, ins)$ \Comment{Extract code and answer by Programmer}
		\State $r = \begin{cases} r, & \text{ success} \\ E, & \text{ error} \end{cases} \gets \text{execute}(C_n)$ \Comment{Code execution, similarly to subsequent $r$}
		
		\While{$r = E$ \textbf{and} $n < T$}  \Comment{Self-correcting mechanism start}
		\State $n \gets n + 1$
		\State $S_n \gets I(C_{n-1}, E)$ \Comment{Inspector provides suggestions}
		\State $C_n \gets A_n, A_n \gets Pr(C_{n-1}, S_n, E)$ \Comment{Programmer modifies code}
		\State $r \gets \text{execute}(C_n)$ \Comment{Execute modified code}
		\EndWhile
		\If{$r = E$}
		\State $r \gets \text{execute}(C_h)$ \Comment{Human intervention (Optional)}
		\State $R \gets C_h \cup Pr(r) $ \Comment{Final response in natural language}
		\EndIf
		\State $R \gets C_n \cup Pr(r) $ \Comment{Final response in natural language}
	\end{algorithmic}
\end{algorithm}

\subsection{Programmer agent}
The main responsibility of the programmer is to write code and respond to the user. Upon the user's dataset upload, the programmer receives a tailored system prompt that outlines the programmer's role, environmental context, and the I/O formats. This prompt is augmented with examples to facilitate few-shot learning for the programmer. Specifically, the system prompt encompasses the user's working directory, the storage path of the dataset, the dimensions of the dataset, the name of each column, the type of each column, information on missing values, and statistical description.

The programmer's workflow can be summarized as follows: initially, the programmer writes code based on instructions from the user or the inspector; subsequently, the program extracts code blocks from the programmer's output and executes them in the kernel. Finally, the programmer generates a final response based on the execution results and communicates it to the user. This final response consists of a summary and suggestions for the next steps.

\subsection{Inspector agent and self-correcting mechanism}
The inspector's role is to provide modification suggestions when errors occur in code execution. The prompt of the inspector includes the code written by the programmer during the current dialogue round and the error messages from the kernel. The inspector will offer actionable revision suggestions to the programmer for code correction. This suggestion prompt contains the erroneous code, kernel error messages, and the inspector's suggestions. This collaborative process between the two agents iterates several rounds until the code executes successfully or the maximum number of attempts is reached. This self-correcting mechanism enables the programmer and inspector to make multiple attempts in case of error. A case of self-correcting mechanism and released experiment can be found in the Supplementary Materials.

\subsection{Integrating human intelligence and AI}
Beyond leveraging the inherent knowledge of LLMs, LAMBDA is further enhanced to integrate human intelligence through external resources such as customized algorithms and statistical models from users. As mentioned above, the challenges faced by general RAG methods in data analysis stem from the potential lack of clear correlation between user instructions and code fragments in the representation space, as well as the impact of the length of code fragments. We design a special KV knowledge base for this challenge.

\begin{figure}[H]
	\begin{center}
		\includegraphics[width=0.9\textwidth]{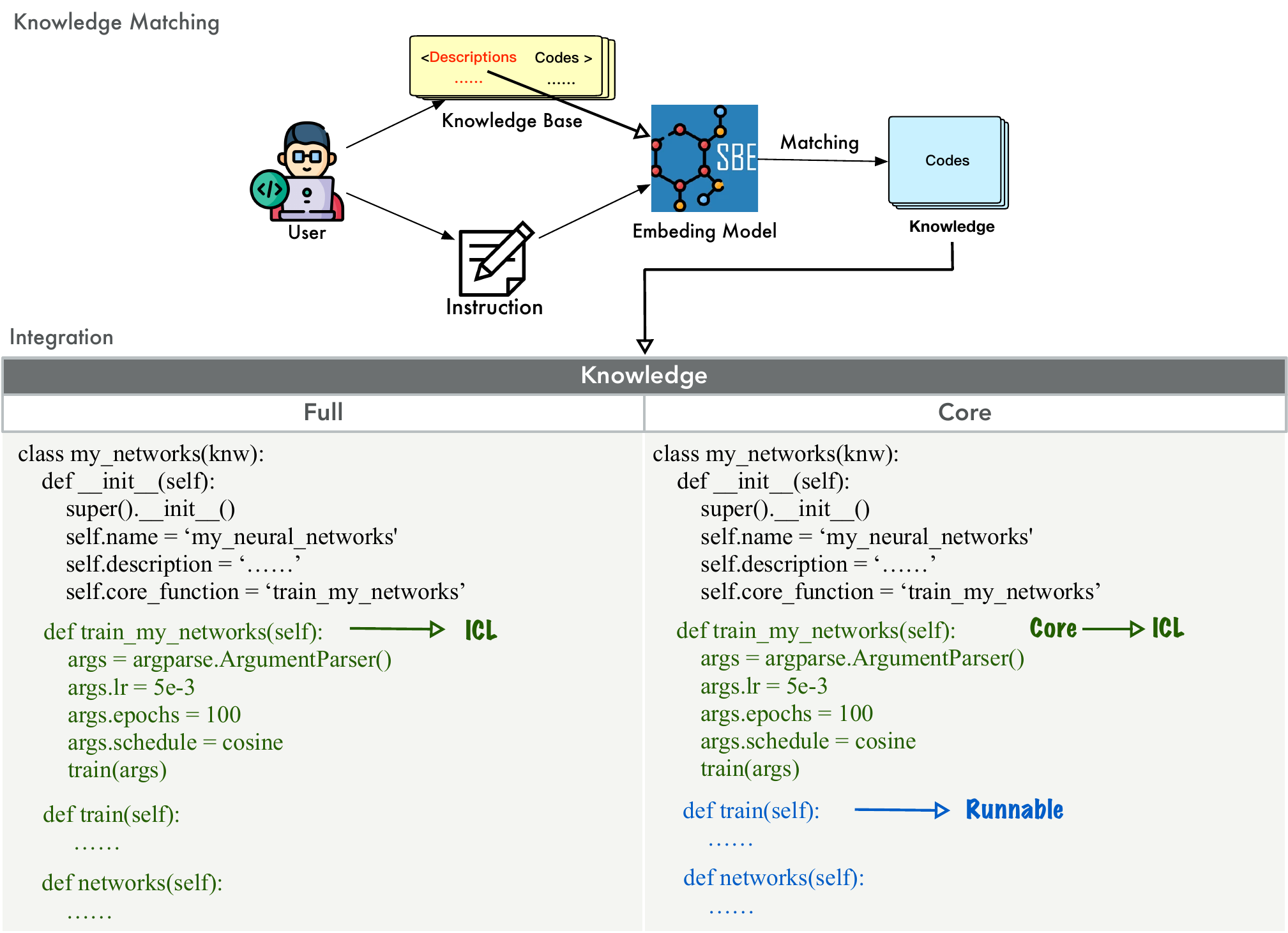}
	\end{center}
	\caption{Knowledge Integration Mechanism in LAMBDA: Knowledge Matching selects codes from the knowledge base by comparing descriptions with the instruction. Two integration modes are available: `Full' mode injects the entire knowledge code into the LLM via ICL, while `Core' mode segments the code into essential usage for ICL and runnable code for back-end execution.
\label{fig:knw}}
\end{figure}

The KV knowledge base is a repository for housing external resources from users in key and value pairs. Specifically, we format the code of resources into key-value pairs: the key represents the resource description, and the value denotes the code. The user's query will be matched within the knowledge base to select the code with the highest similarity. Figure \ref{fig:knw} demonstrates the workflow of knowledge matching in LAMBDA. We define the knowledge base as
$
\mathcal{K} = \{(d_i, c_i) \mid i = 1, 2, \ldots, n\},
$
where \( d_i \) represents the description of the \( i \)-th piece of knowledge and \( c_i \) represents the corresponding source code.

When the user issues an instruction \( ins \), an embedding model \( \mathcal{F} \) encodes all descriptions in the knowledge base and the \(ins \), such as Sentence-BERT \citep{reimers-2019-sentence-bert}. The embedding tensors for descriptions and instruction are represented by \( \mathbf{e}_{d_i} \) and \( \mathbf{e}_{ins} \) respectively. The cosine similarity between them  is calculated to select knowledge with a similarity score greater than a threshold \( \theta \), with the highest-scoring match chosen as the relevant knowledge. 

Let the embedding function be \( \mathcal{F} \), the \( \mathbf{e}_{d_i} \) and \( \mathbf{e}_{ins} \) are formulated as follows
$
\mathbf{e}_{d_i} = \mathcal{F}(d_i), i \in \{1, 2, \ldots, n\},$ and  $\mathbf{e}_{ins} = \mathcal{F}(ins).$
The similarity \( S_i \) between description and instruction is computed using cosine similarity as
\[
S_i(\mathbf{e}_{d_i}, \mathbf{e}_{ins}) = \frac{\mathbf{e}_{d_i} \cdot \mathbf{e}_{ins}}{\|\mathbf{e}_{d_i}\| \|\mathbf{e}_{ins}\|} \quad \forall i \in \{1, 2, \ldots, n\}.
\]

The matched knowledge $k$ with the highest $S_i$ is selected while it satisfies $S_i > \theta$, computed as
\[
k = c_{i^*}, \quad i^* = \arg\max_{i} \left( S_i(\mathbf{e}_{d_i}, \mathbf{e}_{ins}) \cdot \mathbf{1}_{\{S_i(\mathbf{e}_{d_i}, \mathbf{e}_{ins}) > \theta\}} \right) \quad \forall i \in \{1, 2, \ldots, n\}.
\]
The knowledge $k$ will be embedded in ICL for the LLM to generate answer \( \hat{A} \). Formally, given a query \( q \), matched knowledge $k$, a set of demonstrations
\( D = \{(q_1, k_1, a_1), \allowbreak (q_2, k_2, a_2), \allowbreak \ldots, \allowbreak (q_n, k_n, a_n)\}\),
 and the LLM \( \mathcal{M} \), the model estimates the probability \( \mathcal{P}(a|q, k, D) \) and outputs the answer \( \hat{A} \) that maximizes this probability. The final response \( \hat{A} \) is generated by the model \( \mathcal{M} \) as
$
\hat{A} \gets \mathcal{M}(q, D).
$

The matching threshold \( \theta \) defines the required similarity between a knowledge description and a user instruction,  directly influencing the complexity of retrieving relevant knowledge. A higher \( \theta \) imposes stricter matching criteria, reducing the chance of retrieval, whereas a lower \( \theta \) increases the probability of identifying a match.

The optimal selection of \( \theta \) depends on multiple factors. For example, when users aim to incorporate specific knowledge into a task, a lower \( \theta \) value increases the chance of retrieving the relevant information. Furthermore, the length of the knowledge description plays a critical role, as longer descriptions typically necessitate a lower \( \theta \) value since user instructions are generally more concise. By default, we recommend setting \( \theta \) to 0.2. However, this value can be adjusted based on the aforementioned factors to optimize retrieval performance.

By integrating \( k \) through ICL, the model effectively combines retrieved domain knowledge with contextual learning to provide answers that are more accurate. Moreover, LAMBDA offers two integration modes: `Full' and `Core'. In the `Full' mode, the entire knowledge is utilized as the context in ICL. In the `Core' mode, the core functions are processed through ICL, while other functions are executed directly in the back-end. This approach allows the agents to focus on modifying the core function directly, without the need to understand or implement the sub-functions within it. The `Core' mode is particularly effective for scenarios involving lengthy code, as it eliminates the need to process the entire code through ICL. These two modes of knowledge integration provide substantial flexibility for handling tasks that require domain-specific knowledge. We evaluate our Knowledge Integration Mechanism in Table \ref{exp:knw} through several domain tasks.

In summary, the Knowledge Integration Mechanism empowers LAMBDA to perform domain tasks and offers the flexibility needed to address complex data analysis challenges.

\subsection{Kernel, report generation and code exporting}
LAMBDA uses IPython as its kernel to manage sequential data processing, where each operation builds on the previous one, such as standardization followed by one-hot encoding. Implementation details are in the Supplementary Materials. LAMBDA also generates analysis reports from dialogue history, including data processing steps, visualizations, model descriptions, and evaluation results. Users can choose from various report templates, and the agent creates reports via ICL, allowing users to focus on higher-value tasks. A sample report is in Figure \ref{fig:demo} and the Supplementary Materials. Moreover, users can download their experimental code as an IPython notebook.

\subsection{User interface}
LAMBDA provides an accessible user experience similar to ChatGPT. Users can upload datasets and describe tasks in natural language, supported by LLMs like Qwen-2, which recognizes 27 languages. It is recommended to prompt LAMBDA step-by-step, mimicking data analysts' approach, to maintain control and embody the ``human-in-the-loop" concept. LAMBDA generates results, including code, figures, and models, which users can copy and save with a single click. Even those without expertise in statistics or data science can train advanced models by simply asking for recommendations, such as XGBoost and AdaBoost. Advanced users can customize LAMBDA’s knowledge via an interface template. Users can also export text reports and code for further study. A usage example is shown in Figure \ref{fig:demo}. LAMBDA’s interface is designed to be accessible to users of all backgrounds.

To summarize, the programmer agent, inspector agent, self-correcting mechanism, and human-in-the-loop components collectively ensure the reliability of LAMBDA. The integration of knowledge makes LAMBDA scalable and flexible for domain-specific tasks. To enhance portability, we provide an OpenAI-style interface for LAMBDA. This ensures that most LLMs, once deployed via open-source frameworks such as vLLM \citep{10.1145/3600006.3613165} and LLaMA-Factory \citep{zheng2024llamafactory}, are compatible with LAMBDA.

\subsection{Prompt}
We present examples of prompts for the roles of programmer, inspector, self-corrector, and knowledge integrator. Additional prompt examples and case studies are available in the Supplementary Materials.

Figure \ref{fig:pmt-programmer} gives an example prompt for the data analyst at the start of the analysis session.

\begin{figure}[H]
\centering
		\includegraphics[width=\textwidth]{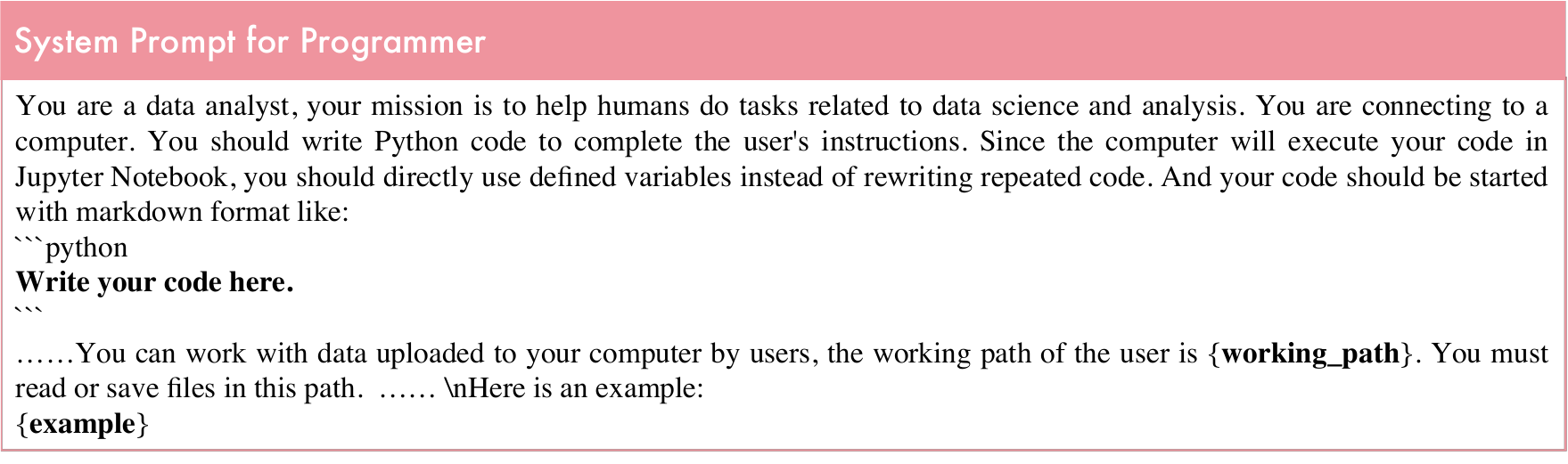}
	\caption{Prompt example for the data analyst.\label{fig:pmt-programmer}}
\end{figure}
\begin{figure}[H]
\centering
		\includegraphics[width=\textwidth]{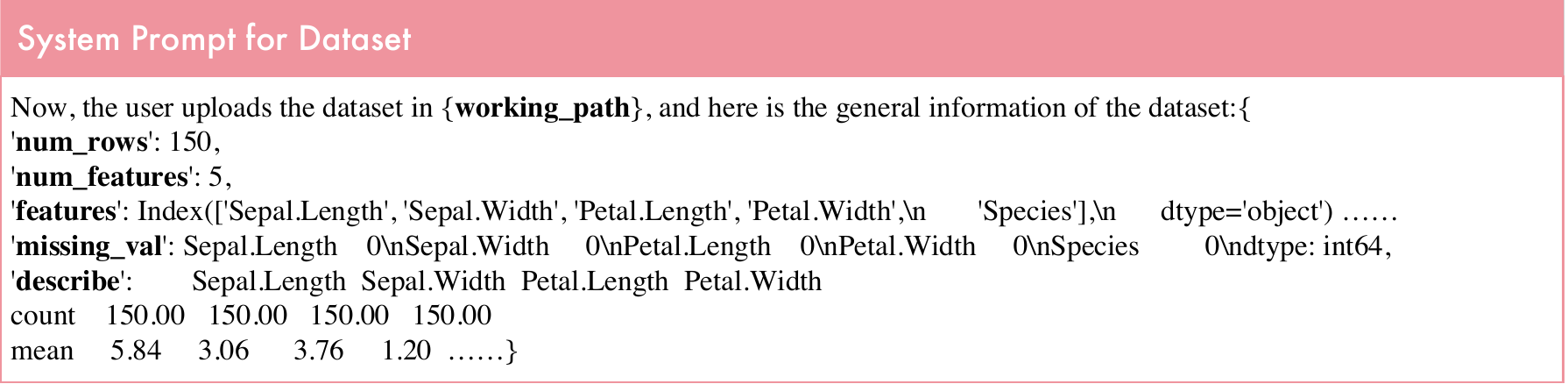}
	\caption{Prompt example for the dataset.\label{fig:pmt-data}}
\end{figure}

Figure \ref{fig:pmt-data} shows a system prompt about the dataset, which provides essential information to the programmer agent.

\begin{figure}[H]
\centering
		\includegraphics[width=\textwidth]{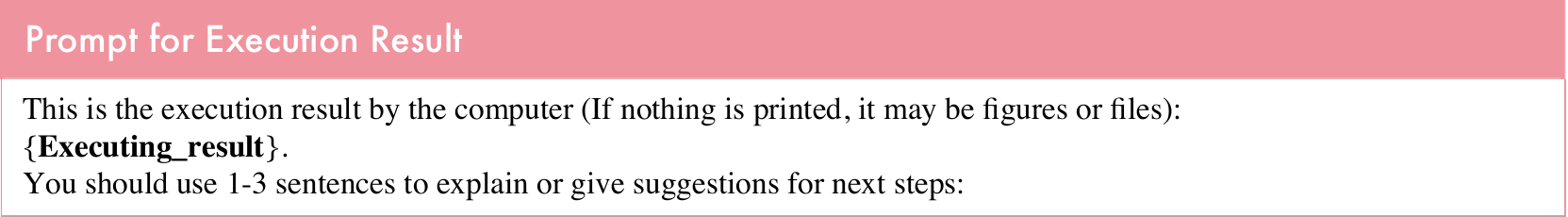}
	\caption{Prompt example for the execution result.\label{fig:pmt-result}}
\end{figure}

After obtaining the execution results, a prompt such as the one given in Figure \ref{fig:pmt-result} can be used to format the output, enabling the programmer agent to provide an explanation or suggest the next steps.

\begin{figure}[H]
\centering
		\includegraphics[width=\textwidth]{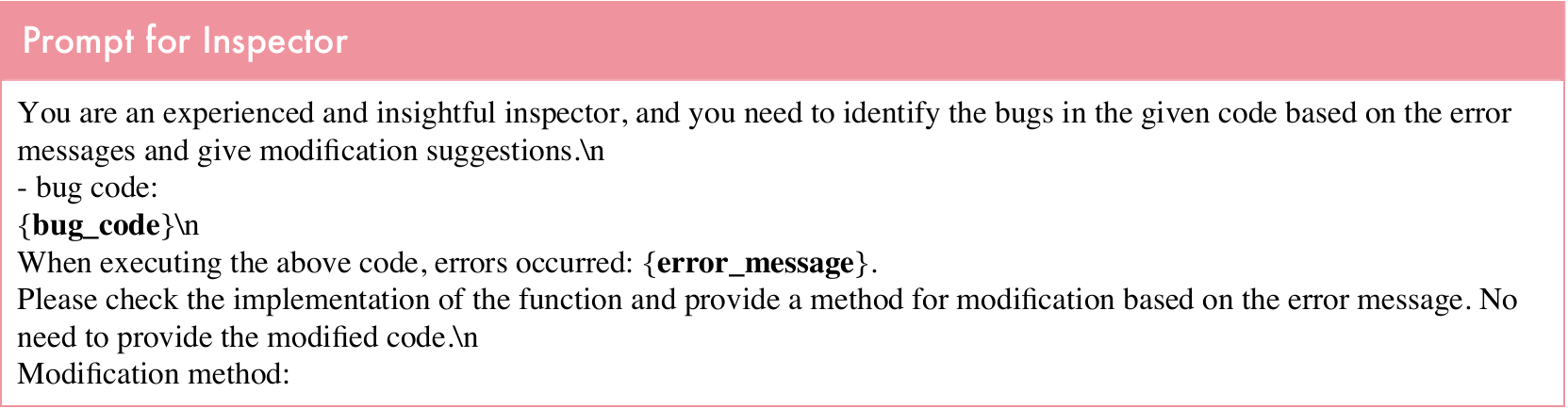}
	\caption{Prompt example for inspector.\label{fig:pmt-inspector}}
\end{figure}

When an error occurs, a prompt for the inspector is employed to guide the inspector in identifying the cause of the bug and to offer revision suggestions (Figure \ref{fig:pmt-inspector}).

\begin{figure}[H]
\centering
		\includegraphics[width=\textwidth]{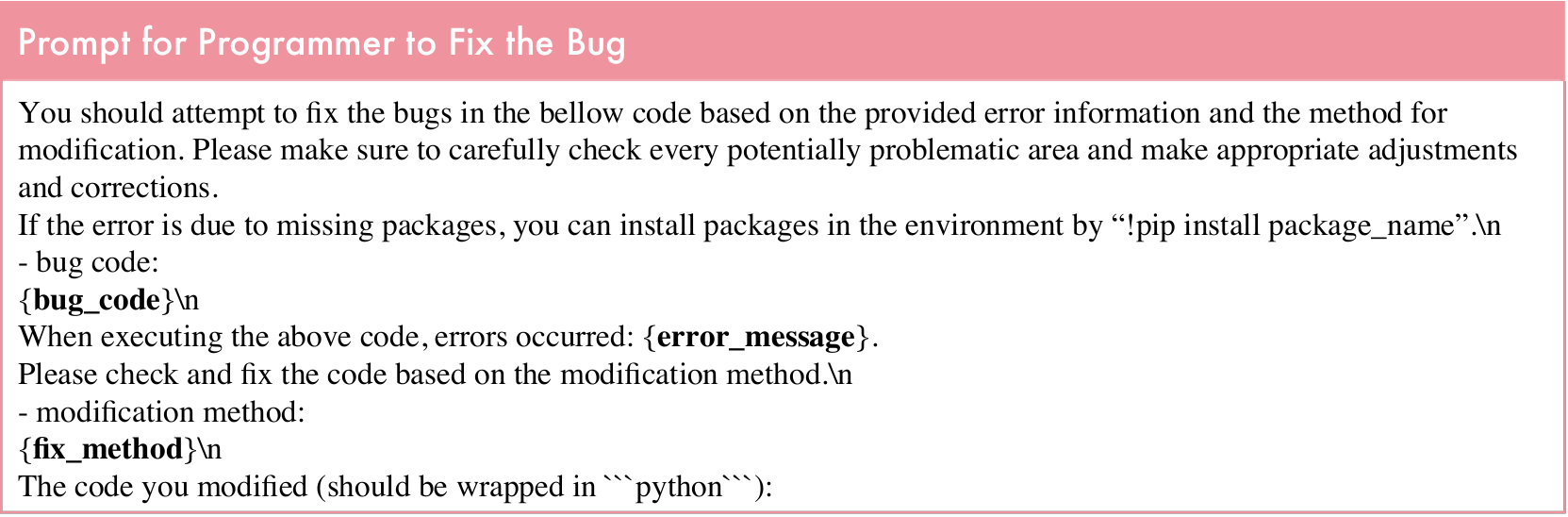}
	\caption{Prompt example for code correction.\label{fig:pmt-fix}}
\end{figure}

Figure \ref{fig:pmt-fix} presents an example prompt for the programmer revising the error code.

\begin{figure}[H]
\centering
		\includegraphics[width=\textwidth]{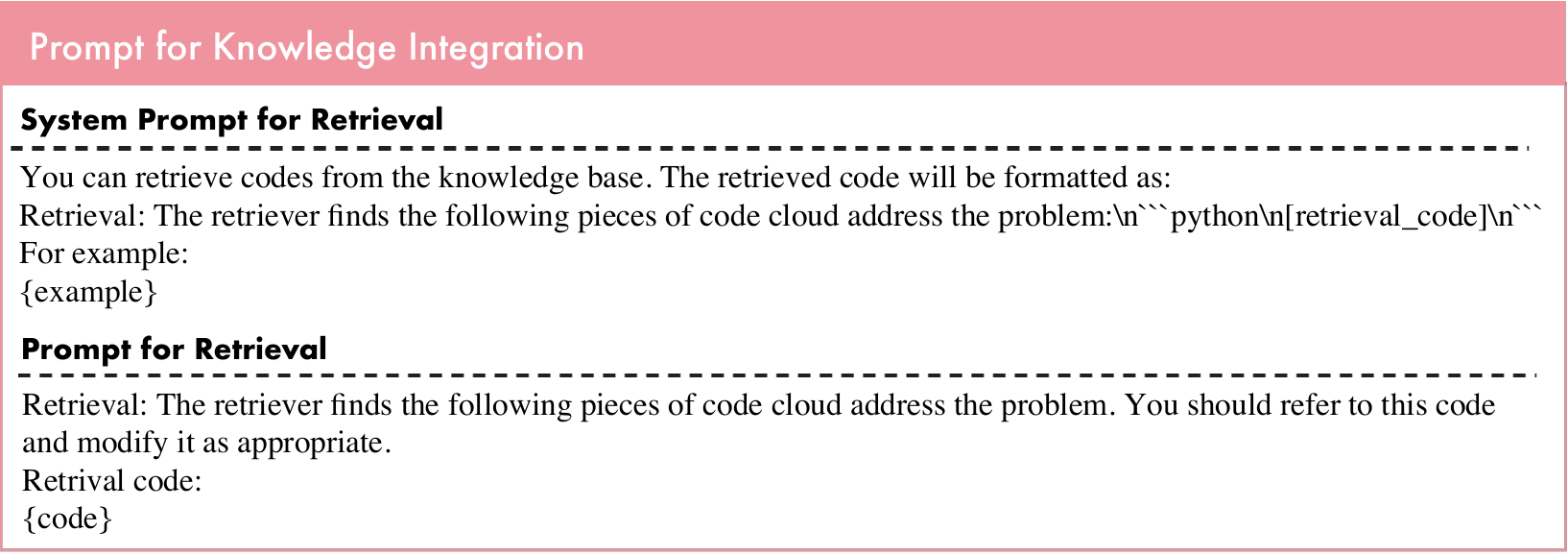}
	\caption{Prompt example for knowledge integration.\label{fig:pmt-knw}}
\end{figure}

For knowledge integration, the system message prompt and retrieval result are shown in Figure \ref{fig:pmt-knw}.

\section{Experiments}
\label{sec:exp_result}

\subsection{Data experiments}
The current data analysis paradigm relies on programming software and languages such as R \citep{R}, SAS \citep{SAS}, and Python \citep{Python} for computation and experimentation. To gain practical experience and evaluate LAMBDA's performance in real-world data science tasks, we first applied LAMBDA to several standard datasets for classification and regression analysis. In addition, we conducted further investigations in broader statistical analysis scenarios, such as high-dimensional data, missing data, image data, and text data, to examined its robustness and versatility. All information of the datasets used can be found in the supplementary materials.

\begin{table}[H]
	\centering
		\begin{tblr}{
				hline{1,17} = {-}{0.1em},
				hline{2} = {-}{},
			}
			DataSets                                                                                                                                                             & Usage                                     \\
AIDS Clinical Trials Group Study 175 \citep{Hammer1996ATC}     & Classification     \\
NHANES \citep{misc_national_health_and_nutrition_health_survey_2013-2014_(nhanes)_age_prediction_subset_887}.
	& Classification    \\
	Breast Cancer Wisconsin \citep{misc_breast_cancer_wisconsin_(diagnostic)_17}  & Classification      \\
			Wine \citep{misc_wine_109}    & Classification       \\
			Concrete Compressive Strength  \citep{misc_concrete_compressive_strength_165}      & Regression       \\
	Combined Cycle Power Plant \citep{misc_combined_cycle_power_plant_294}   & Regression \\
			Abalone \citep{misc_abalone_1}    & Regression - Case Study \\
			Airfoil Self-Noise \citep{misc_airfoil_self-noise_291}  & Regression        \\
			Iris \citep{misc_iris_53}      & Classification - Case Study \\ 
Heart Disease \citep{misc_heart_disease_45} & Regression - Case Study \\
			Genomic Datasets \citep{anhpknu_high_dimensional_data} & High-Dimensional Data \\
			Framingham Heart Study Dataset \citep{framingham_heart_study} & Missing Data \\
			Student Admission Records \citep{kaggle_student_admission} & Missing Data \\
			MINIST \citep{lecun1998mnist} & Image Data \\
			SMS Spam \citep{Almeida2011} & Text Data \\
		\end{tblr}
	\caption{Datasets used in this study.
The Genomic datasets include the following three datasets: TCGAmirna  \citep{bentink2012}, EMTAB386  \citep{colaprico2015}, and GSE49997 \citep{pils2012validating}. }
	\label{table:datasets}
\end{table}

For classification problems, we measured accuracy on the test data, defined as the ratio of correctly classified instances to the total number of instances. For regression problems, we used Mean Squared Error (MSE), which is the average of the squared differences between the predicted values and the actual values in the test data. The formula for MSE is:
$\text{MSE} = (1/n) \sum_{i=1}^{n} (y_i - \hat{y}_i)^2,$
where \( n \) is number of data points, \( y_i \) is the observed value, \( \hat{y}_i \) is the predicted value.  We employed 5-fold cross validation for evaluation in all the cases.
Table \ref{table:datasets} lists the  datasets  used in our experiments and case studies.

\subsubsection{Experiments with classical tabular data}

We initially applied LAMBDA to several classical datasets, covering both classification and regression tasks. To facilitate comparison, we documented the analysis methods employed by LAMBDA and then manually conducted the same analyses using R. The results are summarized in Table \ref{table:reliability-lambda}, with the corresponding results from the R analyses presented in parentheses.

\begin{table}[h]
\centering
\caption{The experimental results obtained using LAMBDA and R are presented, with the R results indicated in parentheses. Classification problems were evaluated using accuracy, where higher values indicate better performance. Regression problems were assessed using mean squared error (MSE), where lower values are preferable. All results were derived from 5-fold cross-validation. The difference result bewteen LAMBDA and R is introduced by different data processing, hyper-paprameters and cross-validation.}
	\label{table:reliability-lambda}
        \begin{adjustbox}{width=\textwidth,totalheight=\textheight,keepaspectratio}
	\begin{tblr}{
			cell{1}{3} = {c=4}{c},
			cell{3}{1} = {r=9}{c},
			cell{14}{1} = {r=9}{},
			hline{1-3,12-14,23-24} = {-}{},
		}
		& \textbf{\textbf{Model}} & \textbf{Datasets} &                      &                  &                    \\
		&                         & \textbf{AIDS }(\%)   & \textbf{NHANES }(\%)     & \textbf{Breast Cancer}(\%)    & \textbf{Wine}(\%) \\
		\textbf{Classification} & Logistic Regression     & 86.54 (86.44)          & 99.43 (99.96)          & \textbf{98.07} (97.72)          & \textbf{98.89} (98.86)           \\
		& SVM                     & 88.45 (88.59)          & 98.82 (98.86)              & 97.72 \textbf{(98.25)}         & \textbf{98.89}  (98.33)           \\
		& Neural Network          & 88.82 (87.89)          & 99.91 (99.91)             & 97.82 (97.01)         & 82.60 (98.87)            \\
		& Decision Tree           & 87.70 (88.78)          & \textbf{100} \textbf{(100)}            & 94.26  (93.32)         & 92.14 (90.91)           \\
		& Random Forest           & 89.29 (88.73)          & \textbf{100} \textbf{(100)}           & 96.84  (95.96)          & 98.33 (98.30)         \\
		& Bagging                 & 89.62 (88.82)          & \textbf{100} \textbf{(100)}          & 96.49  (94.90)          & 96.65 (96.60)         \\
		& Gradient Boost          & 89.20 (88.83)          & \textbf{100} \textbf{(100)}            & 96.84 (94.74)          & 96.65 \textbf{(98.89)}            \\
		& XGBoost                 & \textbf{89.67} (\textbf{89.62})           & \textbf{100} \textbf{(100)}            & 97.54 (97.19)          & 95.54 (98.87)           \\
		& AdaBoost                & 88.92 (89.10)           & \textbf{100}  \textbf{(100)}           & 97.72 (97.55)         & 93.89 (97.71)           \\
		& \textbf{Best Accuracy}          & \textbf{89.67} (89.62)      & \textbf{100} \textbf{(100)}        & 98.07 \textbf{(98.25)}     & \textbf{98.89} \textbf{(98.89)}    \\
&      & \textbf{Concrete} & \textbf{Power Plant} & \textbf{Abalone} & \textbf{Airfoil}   \\
		\textbf{Regression}     & Linear Regression       & 0.4596 (0.3924)             & 0.0714 (0.0713)                & 0.5086 (0.6867)            & 0.5717  (0.6972)              \\
& Lasso  & 0.5609 (0.3918)  & 0.0718 (0.0713) & 0.8042 (0.4739)  & 0.5738 (0.4886)     \\
		& SVR  & 0.4012 (0.4780)  & 0.0534 (0.0489)   &  \textbf{0.4542} \textbf{(0.4408)}            & 0.3854 (0.3725)              \\
		& Neural Network          & \textbf{0.2749} (0.3055)             & 0.0612 (0.0567)                & 0.4551 (0.7185)            & 0.4292 (0.2604)              \\
		& Decision Tree           & 0.5242 (0.5837)              & 0.0551 (0.1175)               & 0.5566 (0.5472)            & 0.3823 (\textbf{0.2559})              \\
		& Random Forest           & 0.4211 (\textbf{0.2755})              & 0.0375 (\textbf{0.0363})                & 0.4749 (0.4460)            & 0.2655 (0.3343)              \\
            & Gradient Boost          & 0.3414 (0.3605)             & \textbf{0.0315} (0.0538)               & 0.4778 (0.5840)             & \textbf{0.2528} (0.2888)               \\
		& XGBoost                 & 0.3221 (0.2991)             & 0.0319 (0.0375)               & 0.4778 (0.4441)            & 0.2741 (0.2832)              \\
		& CatBoost                & 0.2876 (0.4323)            & 0.0325 (0.0568)                & 0.4795 (0.4516)            & 0.2529  (0.2638)             \\
		& \textbf{Best MSE}          & \textbf{0.2749} (0.2755)        & \textbf{0.0315} (0.0363)           & (0.4542) \textbf{0.4408}      & \textbf{0.2528} (0.2559)    \\
	\end{tblr}
    \end{adjustbox}
\end{table}

The results presented in Table \ref{table:reliability-lambda} demonstrate LAMBDA's robust performance in executing data analysis tasks.
These results are either superior to or on par with those obtained using R. These outcomes highlight LAMBDA's effectiveness in leveraging various models across tabular data scenarios. Furthermore, the results indicate that LAMBDA performs at a level comparable to that of a data analyst proficient in R. This suggests the potential for systems like LAMBDA to become indispensable tools for data analysis in the future. Notably, there was no human involvement in the entire experimental process with LAMBDA, as only prompts in English were provided.

In summary, the experimental results demonstrate that LAMBDA achieves human-level performance and can serve as an efficient and reliable data agent, assisting individuals in handling data analysis tasks.

\subsubsection{Experiments with high-dimensional data and unstructured data}
To validate LAMBDA's robustness and versatility, we further explored its application across a broader range of data scenarios, including high-dimensional data, missing data, image data, and text data.

\begin{itemize}
    \item \textbf{High-dimensional data:} We evaluated LAMBDA on the following three challenging high-dimensional clinical datasets: TCGAmirna  \citep{bentink2012}, EMTAB386  \citep{colaprico2015}, and GSE49997 \citep{pils2012validating}.
We summarize the sample size and dimensions  in Table \ref{tb:dm}. The test results are presented in Table \ref{tb:hd}. More detailed descriptions of these three datasets are given in the Supplementary Materials.
We found that LAMBDA consistently applies dimensionality reduction techniques, such as Principal Component Analysis (PCA), as a preprocessing step. This allows us to apply methods like logistic regression without the regularization. The results indicate that LAMBDA is capable of handling high-dimensional data.

    \begin{table}[htbp]
    \centering
    \begin{tblr}{
      hline{1,3} = {-}{0.08em},
      hline{2} = {-}{},
    }
    Data      & TCGAmirna       & EMTAB386       & GSE49997       \\
   {(Size, Dimension)} & (544, 802) & (129, 10360) & (194, 16051) \\
    \end{tblr}
    \caption{Experiment datasets with their sizes and dimensions (rows, columns).}\label{tb:dm}
    \end{table}

    \begin{table}[htbp]
    \centering
    \begin{tblr}{hline{1,11} = {-}{0.08em},
      hline{2} = {-}{},
      hline{10} = {-}{},
    }
    Model              & TCGAmirna (\%) & EMTAB386 (\%) & GSE49997 (\%) \\
    Logistic Regression & 52.58         & 54.18         & 67.52         \\
    Decision Tree       & 54.42         & 57.45         & 63.45         \\
    Random Forest       & 55.16         & 61.20         & 67.54         \\
    Bagging             & \textbf{56.62}         & 58.21         & \textbf{70.63}         \\
    Gradient Boosting   & 54.78         & 55.08         & 70.62         \\
    XGBoost             & 55.15         & 58.15         & 70.62         \\
    AdaBoost            & 55.15         & 57.45         & 70.62         \\
    Neural Network      & 54.22         & \textbf{61.23}         & 66.48         \\
    Best                & \textbf{56.62} & \textbf{61.23} & \textbf{70.63} \\
    \end{tblr}
    \caption{Performance on the high-dimensional datasets. The results are reported in terms of accuracy through 5-fold cross-validation.}\label{tb:hd}
    \end{table}

    \item \textbf{Missing data:} We evaluated LAMBDA on three datasets containing missing values, with results summarized in Table \ref{tb:missing}.
    We observe that LAMBDA tends to prioritize deleting the observations that contain missing values. However, with an appropriate prompt, LAMBDA can also attempt to impute missing values (e.g., mean value). When errors arise due to missing values, the Inspector agent effectively identifies the issue, notifies the Programmer agent, and applies the necessary corrections.

    \begin{table}[htbp]
    \centering
    \begin{tblr}{
      hline{1,11} = {-}{0.08em},
      hline{2} = {-}{},
      hline{10} = {-}{},
    }
    Model                 & Framingham (\%) & StuRecord (\%) & Heart Disease (\%)\\
    Logistic Regression   & \textbf{85.35}   & 50.36 & 59.41 \\
    Neural Network        & 84.95          & 57.28 &  \textbf{60.40} \\
    Decision Tree         & 84.27          & 52.96 & 52.49 \\
    Random Forest         & 85.19          & 55.40 & 60.39 \\
    Bagging               & 85.02          & 58.65 & 60.06 \\
    Gradient Boosting     & 85.12          & 60.50 & 58.41 \\
    XGBoost               & 85.19          & \textbf{61.05} & 60.71 \\
    AdaBoost              & 84.98          & 56.63 & 59.42 \\
    Best                  & \textbf{85.35}  & \textbf{61.05} & \textbf{60.40}\\
    \end{tblr}
    \caption{Performance on Framingham, StuRecord and  Heart Disease datasets. The results are reported in terms of accuracy through 5-fold cross-validation.}\label{tb:missing}
\end{table}

    \item \textbf{Image data:}
        We used LAMBDA to train a handwritten digit classifier based on the MNIST dataset. We prompted LAMBDA to utilize various neural network architectures, such as Convolutional Neural Networks (CNNs) and Transformers, as backbone models. The results of this experiment are presented in Table \ref{tb:mnist}. According to Table \ref{tb:mnist}, we find LAMBDA can effectively implement and apply deep learning architectures like CNNs and Transformers for image classification tasks.

    \begin{table}[htbp]
    \centering
    \begin{tblr}{
      hline{1,4} = {-}{0.08em},
      hline{2} = {-}{},
    }
    Model       & Accuracy (\%) \\
    CNN         & 99.19         \\
    Transformer & 97.23         \\
    \end{tblr}
    \caption{Performance on the MNIST Dataset.}
    \label{tb:mnist}
    \end{table}

    \item \textbf{Text data:}
        We used LAMBDA to train a spam detection classifier based on the SMS Spam Collection Dataset. Similar to our approach with image data, we prompted LAMBDA to experiment with different backbone models for this task. The results are summarized in Table \ref{tb:spam}. As shown in Table \ref{tb:spam}, LAMBDA successfully performed text classification tasks. Notably, when prompted to use a Transformer-based architecture, LAMBDA employed DistilBERT-Base-Uncased for transfer learning, which significantly improved both training efficiency and model performance.

    \begin{table}[htbp]
    \centering
    \begin{tblr}{
      hline{1,4} = {-}{0.08em},
      hline{2} = {-}{},
    }
    Model          & Accuracy (\%) \\
    Multinomial Naive Bayes  & 98.39         \\
    BERT           & 99.37         \\
    \end{tblr}
    \caption{Performance of different backbones on the SPAM classification task.}
    \label{tab:model_comparison}\label{tb:spam}
    \end{table}
\end{itemize}

Overall, our findings indicate that LAMBDA is not only capable of handling tabular tabular tasks but also effectively processing image and text data. In future work, we aim to explore more complex and diverse data scenarios.

\subsection{Performance of Knowledge Integration}
We collected three domain-specific tasks to evaluate the proposed Knowledge Integration Mechanism and compare it with advanced data analysis agents. Specifically, the tasks involve utilizing the recent algorithm packages (e.g., PAMI \citep{piotrowski2021fixed}), implementing optimization algorithms (e.g., computing the nearest correlation matrix), and training the latest research models (e.g., non-negative neural networks). For each task, we define a score $\mathcal{S}$ that is calculated as follows:

$$ \mathcal{S} =
    \begin{cases}
    0, & \hspace{0.5em} \text{code error and execution error, or exceeded runtime limit,} \\
    0.5, & \hspace{0.5em} \text{code error and execution successful,} \\
    0.8, & \hspace{0.5em} \text{code successful, execution error due to other issues, e.g. environment,} \\
    1, & \hspace{0.5em} \text{both code and execution successful.}
    \end{cases}
$$

To ensure maximum alignment in experimental settings, we converted the code into corresponding tools for agents equipped with a tools mechanism. For agents lacking such a mechanism, we directly included the code in their context. All agents are implemented using GPT-3.5, except for methods and platforms that have their own models, such as GPT-4-Advanced Data Analysis, ChatGLM-Data Analysis, and OpenCodeInterpreter. Since each task can be completed within one minute, we set a maximum runtime limit of 5 minutes to prevent some agents from becoming stuck in infinite self-modification loops.

\begin{itemize}
    \item \textit{Pattern Mining}\hspace{1em}
        \cite{piotrowski2021fixed} introduce PAMI (PAttern MIning), a cross-platform, open-source Python library offering algorithms to uncover patterns in diverse databases across multiple computing architectures.
    \item \textit{Nearest Correlation Matrix}\hspace{1em} \cite{qincm} propose a Newton-type method specifically designed for the nearest correlation matrix problem. Numerical experiments validate the method’s fast convergence and high efficiency.
    \item \textit{Fixed Points Non-negative Neural Networks}\hspace{1em} \cite{rage2024pami} analyze nonnegative neural networks, which are defined as neural networks that map nonnegative vectors to nonnegative vectors.
\end{itemize}
	
\begin{table}[h]
	\centering
	\begin{tblr}{
		hline{1,10} = {-}{0.08em},
		hline{2} = {-}{},
	  }
							   & PM   & NCM & FPNENN \\
	GPT-4-Advanced Data Analysis \citep{openai2024gpt4}                     & 0.80 (4)  & 0 (1)   & 0 (1)   \\
	ChatGLM-Data Analysis \citep{du2022glm} & 0 (2)    & 0 (2)   & 0 (2)   \\
	OpenInterpreter \citep{openinterpreter}            & 0 (2)    & 0 (2)   & 0 (2)   \\
	OpenCodeInterpreter \citep{zheng2024opencodeinterpreter}        & 1.00 (5)    & 0 (1)   & 0 (1)   \\
	Chapyter \citep{chapyter}             & 0  (2)    & 0 (2)   & 0 (2)   \\
	DataInterpreter (Tools) \citep{hong2024data}            & \textbf{1.00 (5)}  & \textbf{1.00 (5)}   & \textbf{1.00 (5)}   \\
	TaskWeaver (Plugins) \citep{taskweaver}         & \textbf{1.00 (5)}   & \textbf{1.00 (5)}   & \textbf{1.00 (5)}   \\
	\textbf{LAMBDA (Knowledge)}           & \textbf{1.00 (5)} & \textbf{1.00 (5)} & \textbf{1.00 (5)}  \\
	\end{tblr}
	\caption{Performance of the Knowledge Integration Mechanism.
In the table,  `PM' refers to pattern mining, `NCM' refers to the nearest correlation matrix, and `FPNENN' stands for fixed points in non-negative neural networks. The values represent the performance scores, with failure reasons noted in brackets. Specifically, 1: code error and execution error; 2: exceeded runtime limit; 3: code error but successful execution; 4: right code but execution error due to other issues; 5: right code and successful execution.}\label{exp:knw}
\end{table}

Table \ref{exp:knw} demonstrates the effectiveness of LAMBDA's Knowledge Integration mechanism. Specifically, our results showed that many methods scored zero, particularly when the code was lengthy or involved unfamiliar packages not encountered during LLM training. In these situations, most other approaches struggle with one-shot learning. Two exceptions are Data Interpreter and TaskWeaver, which successfully complete the task using pre-defined Plugins/Tools. With the pre-defined Plugins/Tools, they can execute operations internally without requiring the LLM to generate precise code. This mechanism is similar to the `Core' mode of our LAMBDA.

With these tools, the LLM only needs to learn a given code usage example rather than generating the full internal implementation, even when it has access to those details. Although these approaches are generally suitable, the agent is likely to make mistakes when there is the certain misalignment between the users' instructions and integrated knowledge. In such circumstances, we need to utilize the `Full' mode of our LAMBDA.
To further support our claim, we designed two additional experiments.

We take the fixed point non-negative neural networks as a example. We further explore the following two cases that involve misalignment in integrating knowledge/tools and human instruction, which require modifications to the tools (the loss and network mapping are annotated in the schema):

\begin{itemize}
    \item \textbf{Case 1:} The instruction specifies the use of L1 Loss, whereas the tool are originally configured with MSE Loss.
    \item \textbf{Case 2:} The instruction specifies a network structure mapping as follows:
    \begin{itemize}
        \item Encoder: $784 \rightarrow 400$, whereas $784 \rightarrow 200$ originally configured.
        \item Decoder: $400 \rightarrow 784$, whereas $200 \rightarrow 784$ originally configured.
    \end{itemize}
\end{itemize}

\begin{table}[h]
	\centering
	\begin{tabular}{@{}lll@{}}
	\toprule
	Methods               & Misalignment Loss & Misalignment Network \\ \midrule
	TaskWeaver (Plugins)   & \textcolor{red}{\ding{55}} Directly using the plugin      & \textcolor{red}{\ding{55}}
	Directly using the plugin         \\
	Data Interpreter (Tools)     & \textcolor{red}{\ding{55}} Directly use the tool      & \textcolor{red}{\ding{55}} Directly use the tool         \\
	LAMBDA (Knowledge) & \textcolor{teal}{\ding{52}} Alignment      & \textcolor{teal}{\ding{52}} Alignment          \\ \bottomrule
	\end{tabular}
	\caption{The results of case study on Misalignment between Tools and Instructions. Both Plugins and Tools Integration directly use the tools and are not aware of the Misalignment between Tools and Instructions.}\label{case_mis_tools}
\end{table}

From Table \ref{case_mis_tools}, we observe that in Cases 1 and 2, which require modifications to the tools, both TaskWeaver and Data Interpreter directly use the original tools without recognizing that the tools no longer meet the new requirements although the loss and network mapping are annotated in the schema. In contrast, due to the visibility of the knowledge code under `Full' mode, LAMBDA identifies that the original code cannot satisfy the new requirements, makes the necessary adjustments, and successfully completes the two cases.

\section{{Examples}}\label{sec:example}
\begin{figure}[H]
\centering
		\includegraphics[height=0.98\textheight]{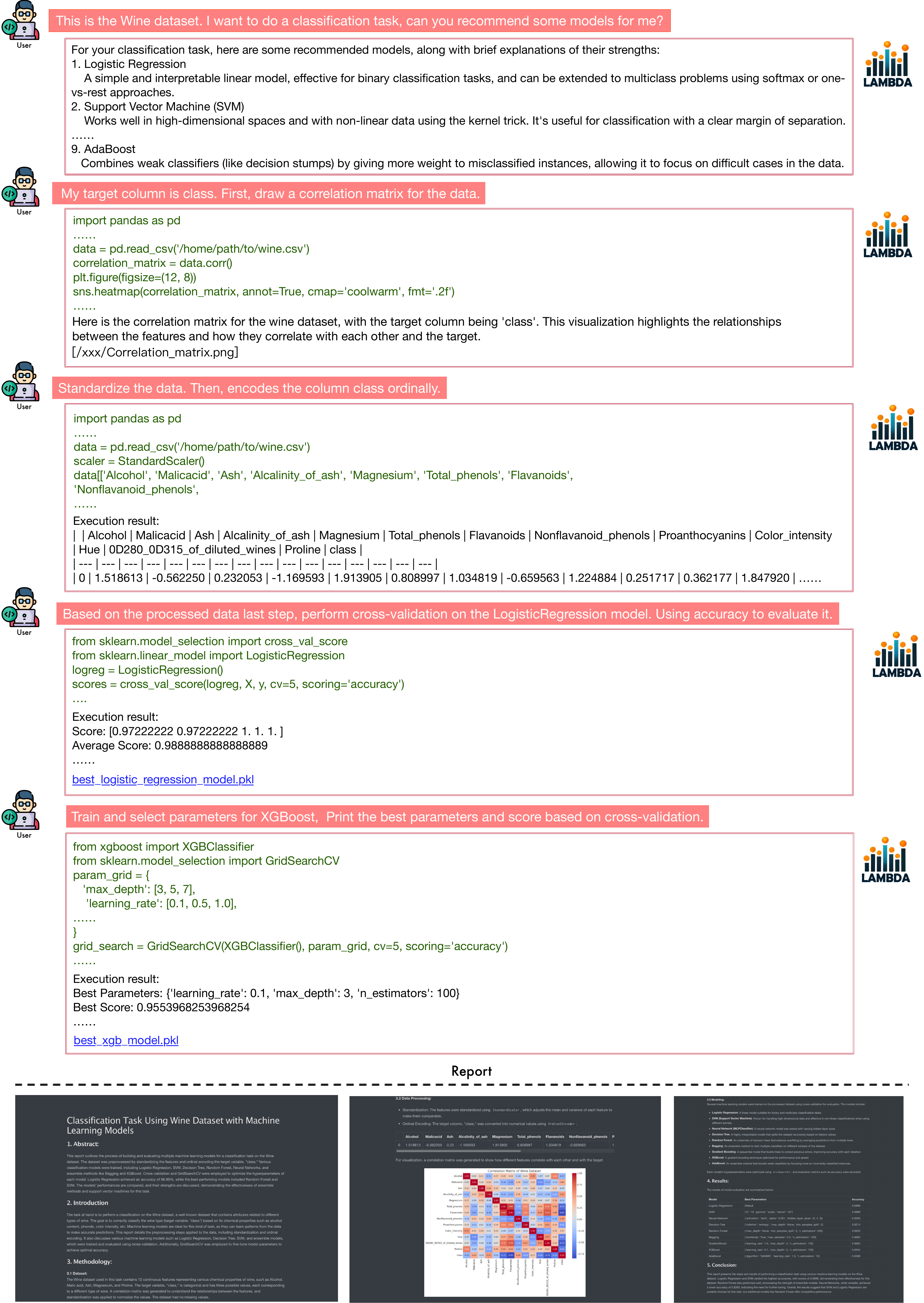}
	\caption{An example of using LAMBDA for classification analysis with the Wine dataset.\label{fig:demo}}
\end{figure}

We present an example of using LAMBDA for building a classification model in Figure \ref{fig:demo}. We also provide three case studies in video format to demonstrate the use of LAMBDA in data analysis, integrating human intelligence and AI, and education.

\vspace{-0.1in}
\begin{itemize}
    \item \textit{Data Analysis}\hspace{1em} We simulate scenarios in which the user requests LAMBDA to perform various tasks, including data preprocessing, data visualization, and model training, on the provided Iris dataset \citep{misc_iris_53}. LAMBDA consistently delivers accurate responses. Additionally, LAMBDA generates an analysis report based on the chat history. A demonstration of this process is given in the first video at
        \url{https://www.polyu.edu.hk/ama/cmfai/lambda.html}.

\vspace{-0.1in}
    \item \textit{Integrating Human Intelligence and AI}\hspace{1em}
        We demonstrated the Knowledge Integration capabilities of LAMBDA by computing the nearest correlation matrix using the Quadratically Convergent Newton Method. We first highlighted the limitations of GPT-4-Advanced Data Analysis in performing this task, thereby underscoring the value of LAMBDA through comparison. A demonstration is given in the second video at
        \url{https://www.polyu.edu.hk/ama/cmfai/lambda.html}.

\vspace{-0.1in}
    \item \textit{Interactive Education}\hspace{1em}
    We consider an educational scenario in which the teacher uses LAMBDA to design the exercise assignments, and the students use LAMBDA to complete exercises. The exercise dataset used is Abalone. This educational support system enhances the efficiency of both teaching and learning. A demonstration is given in the third video at \url{https://www.polyu.edu.hk/ama/cmfai/lambda.html}.
\end{itemize}

\section{Conclusion}\label{sec:con}

LAMBDA is an open-source multi-agent data analysis system that effectively integrates human intelligence with artificial intelligence. Experimental results demonstrate that LAMBDA  achieves satisfactory performance in handling various data analysis tasks. In the future, LAMBDA can be further enhanced with advanced planning, reasoning techniques, and knowledge integration methods to address a broader range of domain-specific tasks. Our results and examples underscore the significant potential of LAMBDA to enhance both statistical and data science practice and education.

By bridging the gap between human expertise and AI capabilities, LAMBDA aims to democratize data science and statistical analysis, fostering a more inclusive environment for innovation and discovery. Its open-source nature encourages collaboration and continuous improvement from the global research community, allowing researchers and developers to contribute to its evolution. As LAMBDA continues to develop, it has the potential to become an invaluable tool for statisticians, data scientists, and domain experts, enhancing their ability to analyze data efficiently and effectively.

Moreover, LAMBDA holds significant potential for statistical and data science education. Its natural language interface lowers barriers for educators and students, enabling them to focus on problem formulation rather than getting bogged down by syntactic complexities. By generating executable code for various tasks, LAMBDA provides immediate, actionable feedback, which can enhance the learning experience by allowing students to see the direct impact of their queries and hypotheses. This capability not only aids in teaching fundamental concepts but also empowers students to experiment and explore data-driven insights independently.

Future work on LAMBDA could focus on several key areas. First, enhancing LAMBDA's ability to seamlessly integrate and leverage large models from various domains for statistical analysis could significantly improve its capacity to tackle complex data analysis tasks. Second, improving the user interface and increasing user satisfaction would make the system more accessible to non-experts. Third, incorporating real-time data processing capabilities could enable LAMBDA to handle streaming data, which is increasingly important in many applications. Finally, expanding the system's support for collaborative work among multiple users could further enhance its utility in both educational and professional settings. We plan to implement LAMBDA in our classroom teaching scenarios, continuously gather feedback from various groups, and use user satisfaction as a metric for evaluating LAMBDA.

In conclusion, LAMBDA represents a meaningful step forward in integrating human and artificial intelligence for data analysis. Its continued development and refinement have the potential to advance the fields of statistics and data science, making sophisticated analytical tools more accessible to users from diverse backgrounds. We have made our code available at
\url{https://github.com/AMA-CMFAI/LAMBDA}.

\subsection*{Acknowledgments}
This work was funded by the Centre for the Mathematical Foundations of Generative AI
and the research grants from The Hong Kong Polytechnic University (P0046811).
The research of Ruijian Han was partially supported by The Hong Kong Polytechnic University (P0044617, P0045351, P0050935).
The research of Houduo Qi was partially supported by the Hong Kong RGC grant (15309223) and
The Hong Kong Polytechnic University (P0045347).
The research of Defeng Sun and Yancheng Yuan was partially supported by the Research Center for Intelligent Operations Research at The Hong Kong Polytechnic University (P0051214).
The research of Jian Huang was partially supported by The Hong Kong Polytechnic University (P0042888, P0045417, P0045931).

{\singlespace
\bibliographystyle{apalike}
\bibliography{lambda3.bib} 
}

\appendix
\setcounter{equation}{0}  
\renewcommand{\theequation}{S.\arabic{equation}}
\setcounter{table}{0}
\renewcommand{\thetable}{S.\arabic{table}}
\setcounter{figure}{0}
\renewcommand{\thefigure}{S.\arabic{figure}}
\setcounter{equation}{0}  

\newpage
\begin{center}
\textbf{\Large Supplementary Materials}
\end{center}

The Supplementary Materials provide key methodological details, experimental data, case studies, and the experimental setup. Specifically, Section \ref{init_idea} introduces our initial concept of a function-calling-based agent system along with relevant discussions. Section \ref{kernel} details the core modules involved in kernel development and describes the datasets used in our experiments. Section \ref{knowledge} elaborates on the implementation of the Knowledge Base and Knowledge Integration. In Section \ref{disc}, we analyze the distinctions between LAMBDA and related approaches, including MetaGPT and ChatGPT-Advanced Data Analysis. Section \ref{data} presents additional details on the datasets, while Section \ref{case_study} presents multiple case studies, demonstrating LAMBDA’s capabilities in data analysis, self-correction mechanisms, human intelligence integration, educational applications, and report generation. Finally, Section \ref{exp_set} outlines the experimental setup.

\section{Function Calling Based Agent System \label{init_idea}}
\begin{figure}[h]
	\begin{center}
		\includegraphics[width=\textwidth]{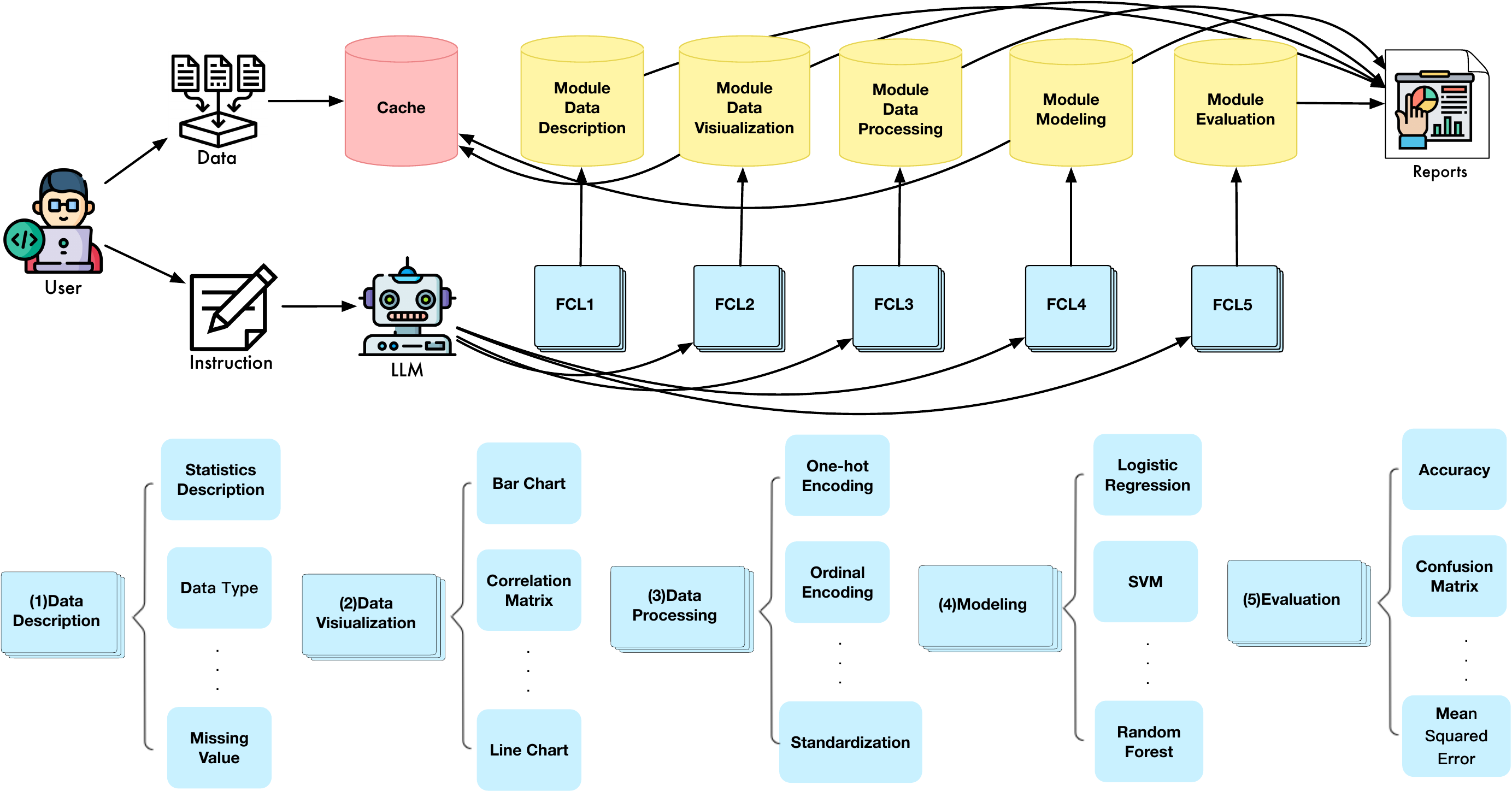}
	\end{center}
	\caption{Agent system design by the function calling method. The FCL means function calling library. \label{fig:fc}}
\end{figure}

Our initial idea was to implement function calling. We developed extensive APIs that encompass a wide range of data processing and machine learning functionalities, including statistical descriptions (e.g., mean, median, standard deviation), encoding schemes (e.g., one-hot encoding, ordinal encoding), data partitioning, and model training (e.g., logistic regression, decision tree). We utilized five function libraries to build these APIs, each tailored for different purposes: the Data Description Library, Data Visualization Library, Data Processing Library, Modeling Library, and Evaluation Library. Each library caches variables such as processed data and models throughout the program's lifecycle. The framework and workflow are illustrated in Figure \ref{fig:fc}.

\begin{figure}[h]
	\begin{center}
		\includegraphics[width=\textwidth]{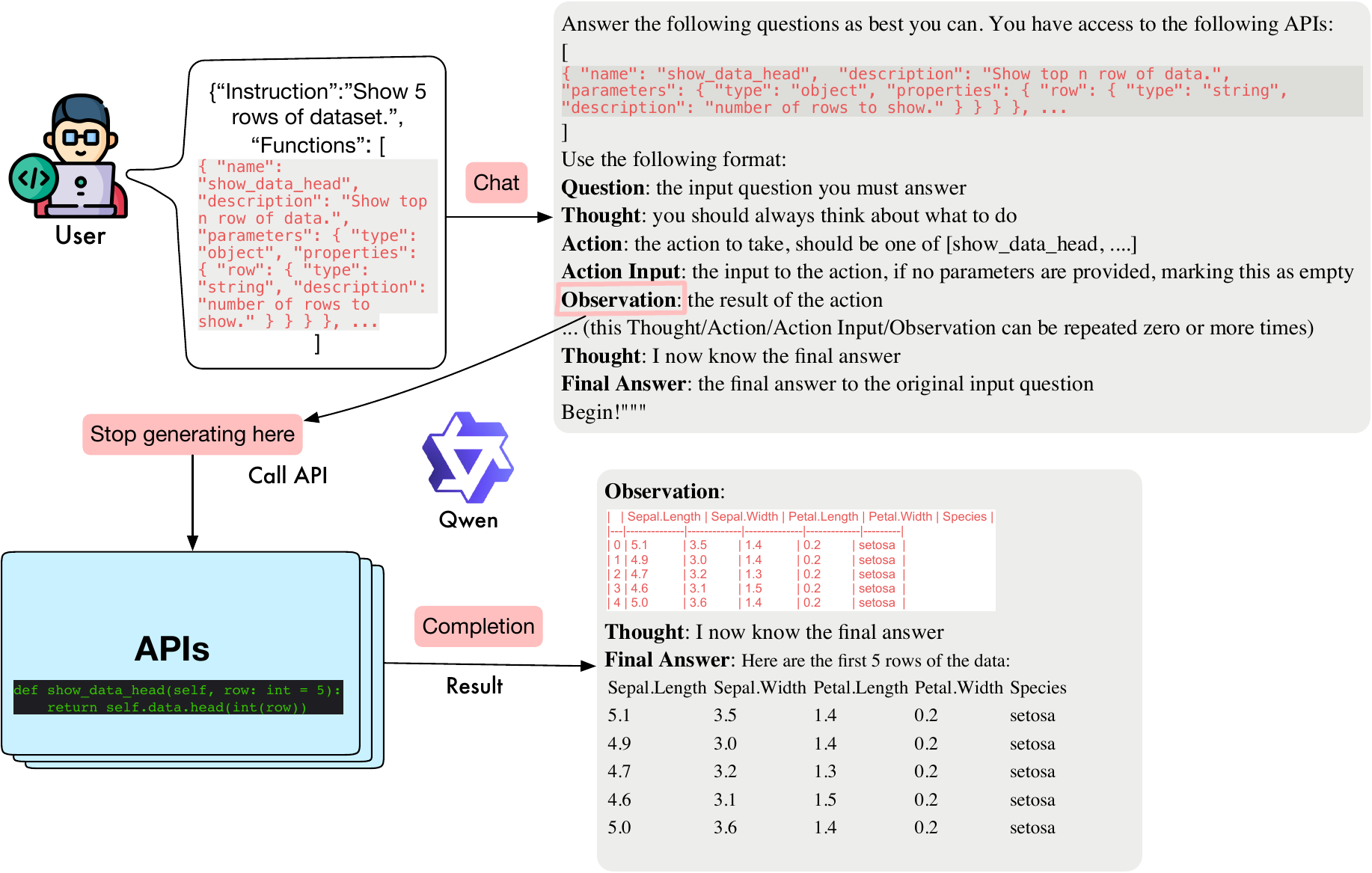}
	\end{center}
	\caption{Workflow of function calling service, demonstrated by Qwen1.5 and ReAct.\label{fig:qwen}}
\end{figure}

We implemented the function calling service by ReAct. Specifically, when prompted to generate text up to the ``Observation"  section, the LLM should halt generation at this point. This is essential as the ``Observation" section requires the outcome of API execution to prevent LLMs from generating results autonomously. The details are depicted in Figure \ref{fig:qwen}.

However, during the implementation, we found the function calling based data agent have some drawbacks and are engineering impractical.

\begin{itemize}\label{conj}
	\item The function-calling method requires APIs to be pre-defined in advance. It is easy to imagine that the number of APIs required for statistical analysis and data processing across various scenarios is virtually limitless, making this approach impractical.
	\item In applications, a large number of APIs are typically required. These APIs often have complex interrelationships and extensive annotations. Such lengthy API annotations can result in sequences that exceed the maximum processing capacity of current LLMs, leading to the truncation of context.
	\item The model's ability to accurately select APIs diminishes as the number of available APIs increases. This decline is due to the increased complexity introduced by the growing number of APIs that LLMs need to evaluate. An incorrect choice of tools or models can directly lead to erroneous results and answers.
\end{itemize}

Building upon these conjections, we have designed some experiments to verify it.

\subsection{Challenges of the function-calling method}\label{result:comparsion}
\begin{figure}[H]
\centering
		\includegraphics[width=0.8\textwidth]{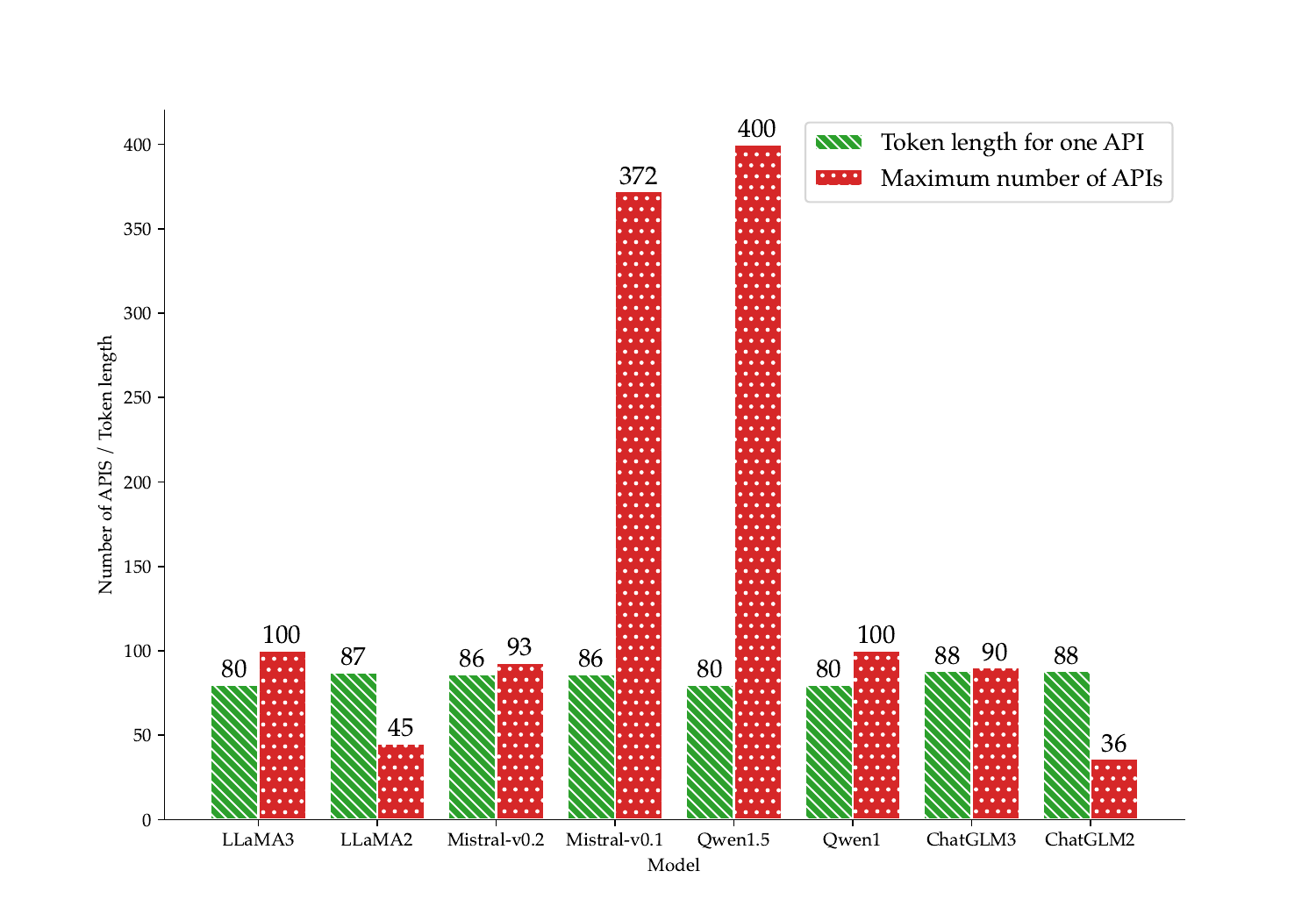}
	\caption{Average token length for one API and maximum number of APIs each LLM can process. \label{fig:max}}
\end{figure}

We estimate the maximum number of APIs that some open-source LLMs can handle in data analysis by using the average length of the pre-defined APIs. Figure \ref{fig:max} illustrates the results. Qwen1.5 and Mistral-v0.1 were specifically designed to handle lengthy sequences, capable of managing 400 and 372 APIs, respectively. However, general-purpose LLMs such as LLaMA2, LLaMA3, Mistral-V0.2, Qwen1, ChatGLM2, and ChatGLM3 can process fewer than 100 APIs. This limitation poses a challenge for applications requiring a larger number of APIs, such as data analysis tasks.

\begin{figure}[h]
	\centering
	\includegraphics[width=0.7\textwidth]{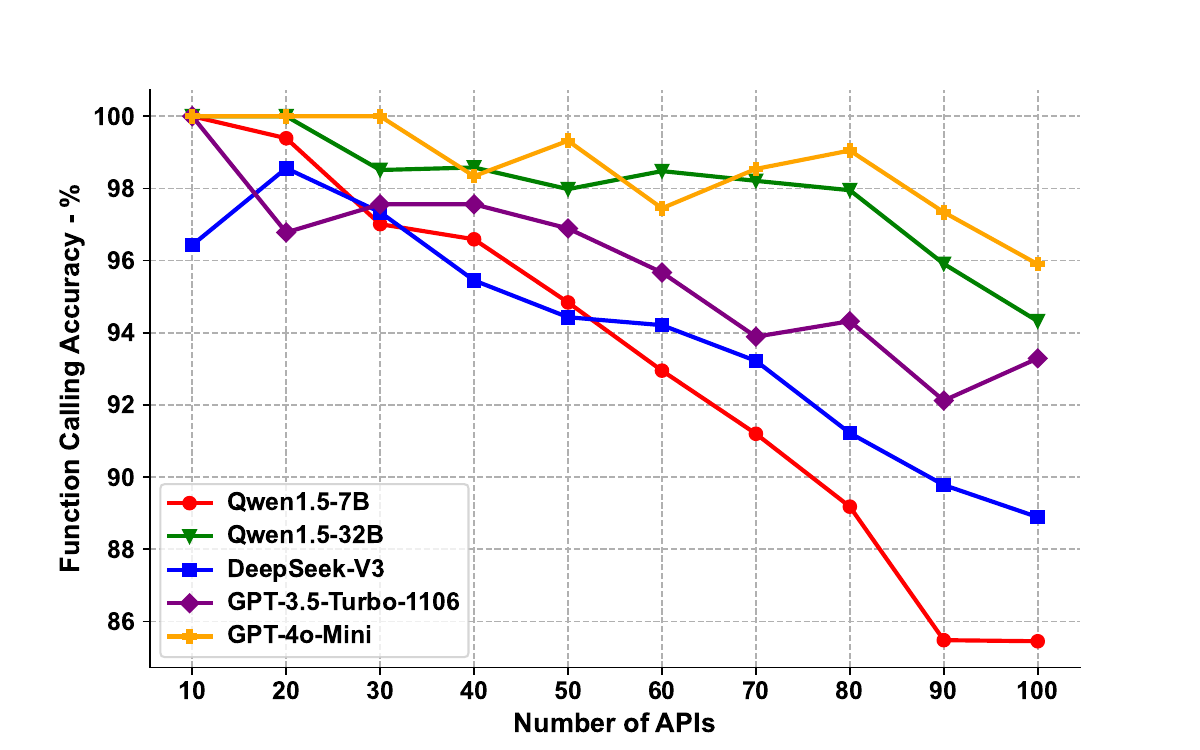}
	\caption{The accuracy of API chosen by models Qwen, DeepSeek, and GPT series.\label{fig:fca}}
\end{figure}

To investigate the impact of the number of APIs on the accuracy of LLMs in selecting the correct APIs, we constructed a dataset comprising 100 commonly used APIs in data analysis. Using few-shot learning, we generated 880 testing instructions aligned with these APIs using Qwen1.5-110B. We then segmented both the APIs and testing instructions into intervals of 10 functions for analysis. The details of the evaluation dataset are shown in Table \ref{table:fca}, and the results are presented in Figure \ref{fig:fca}.

\begin{table}[H]
	\centering
	\caption{Number of APIs and corresponding instructions in the evaluation dataset. \label{table:fca}}
	\begin{tabular}{@{}lllllllllll@{}}
		\toprule
		& \multicolumn{10}{c}{APIs}                           \\
		& 10 & 20  & 30  & 40  & 50  & 60  & 70  & 80  & 90  & 100 \\ \midrule
		\multicolumn{1}{c}{Instructions} & 74 & 163 & 268 & 352 & 446 & 525 & 614 & 684 & 806 & 880 \\ \bottomrule
	\end{tabular}
\end{table}


The results of API selection indicate a notable decline in the model's ability to accurately select APIs as the number of APIs increases.
In the data science scenario, the number of APIs can be extensive due to it encompassing various processing methods and combinations.

In summary, the function-calling method exhibits several significant drawbacks. Firstly, the labor-intensive process of defining numerous APIs is engineeringly impractical. Secondly, the static nature of APIs hinders adaptability to diverse and evolving user demands. Thirdly, extensive API annotations can occupy a substantial portion of the input sequence, potentially leading to truncation risks. Lastly, as the number of APIs increases, the model's accuracy in correct selection decreases, thereby introducing potential inaccuracies in results.

\subsection{Evaluation of the inspector agent}
To assess the importance of the inspector agent in LAMBDA, we designed an ablation study using the heart disease dataset. This dataset, which contains missing values, naturally presents challenges. We utilized Qwen1.5-110B to generate instructions for related tasks. After filtering, there were 454 instructions in the experiment. We evaluated the execution pass rate with only a single agent (programmer)  and two agents (programmer and inspector), respectively.
The execution pass rate is a metric used to evaluate the reliability and effectiveness of a system, process, or code. It is calculated as:
\[
\text{Execution Pass Rate} = \left( \frac{N_{\text{success}}}{N_{\text{total}}} \right) \times 100\%
\]
where \( N_{\text{success}} \) is the number of successful executions and \( N_{\text{total}} \) is the total number of executions attempted. The results are summarized in Table \ref{table:multi-agent}.

\begin{table}[H]
	\caption{Experiment on single agent versus multiple agents. The percentages in brackets are the improvement rate over the single agent. Both the programmer and inspector agent are implemented by Qwen1.5-32b in this experiment.}
	\label{table:multi-agent}
	\centering
	\begin{tabular}{@{}cc@{}}
		\toprule
		Agents                 & Passing Rate \%       \\ \midrule
		programmer agent only  & 68.06          \\
		programmer + inspector & 95.37 (40.13\%) \\ \bottomrule
	\end{tabular}
\end{table}

The result shows a significant gap in the passing rate between using a programmer agent alone and incorporating an inspector. The programmer agent achieved a passing rate of 68.06\%, while the integration of the inspector increased the passing rate to 95.37\%, marking a substantial improvement of 40.13\% over the single-agent setup. This experiment verified the crucial role of collaborative agents in enhancing the reliability and robustness of LAMBDA. By leveraging complementary strengths in error suggestion and correction, the multi-agent collaboration approach not only improves the code passing rate but also reduces the frequency of human intervention in executing data analysis tasks.

\section{Implementation of Kernel}\label{kernel}

The \texttt{CodeKernel} is designed to facilitate the execution of code within a Jupyter Notebook environment. It interacts with the Jupyter backend kernel to manage code execution, handle errors, and oversee the lifecycle of the kernel. This class provides an interface for executing code in a controlled manner, ensuring that outputs are captured and processed effectively.

\subsection{Constructor} 
The constructor of the \texttt{CodeKernel} class initializes an instance of the kernel \texttt{kernel\_name}, \texttt{kernel\_id}, \texttt{kernel\_config\_path}, \texttt{python\_path}, \texttt{ipython\_path}, \texttt{init\_file\_path}, and \texttt{verbose}. The initialization process involves setting up environment variables based on the specified paths for Python or IPython, which are crucial for ensuring the correct execution context.

The kernel is initialized using the \texttt{KernelManager} from the \texttt{jupyter\_client} library, allowing for granular control over the kernel's operations. Depending on whether a configuration file is provided, the kernel is either started with the given settings or defaults are applied. The constructor also provides detailed logging of the kernel's connection information if verbosity is enabled.

\subsection{Code execution} 
The \texttt{execute} method is the primary mechanism for running code within the initialized kernel environment. This method sends a code string to the kernel for execution, utilizing a blocking client to ensure that the process completes before continuing. The method retrieves both shell messages and IOPub messages, which contain critical information about the execution status and outputs.

Outputs are processed to handle standard output (\texttt{stdout}), errors (\texttt{stderr}), and other response types from the kernel. The method returns a tuple consisting of the shell message and a list of processed output messages. This design allows for comprehensive handling of multi-line outputs and ensures that the relevant results are captured and returned.

\subsection{Interactive execution} 
The \texttt{execute\_interactive} method facilitates the execution of code in an interactive manner, where outputs are immediately accessible to the user. This method ensures that the code execution is monitored closely, with specific handling for timeout scenarios. If the \texttt{verbose} flag is set, the method provides detailed output about the execution process, aiding in debugging and analysis.

\subsection{Code inspection} 
The \texttt{inspect} method allows for the introspection of code by sending it to the kernel and retrieving detailed information about variables, functions, and other elements within the code. This method is particularly useful for debugging, as it provides real-time insights into the structure and behavior of the code being executed. The inspection results are returned as part of the shell message, which can be further processed or displayed.

\subsection{Error handling} 
The \texttt{get\_error\_msg} and \texttt{check\_msg} methods are responsible for handling errors that occur during code execution. The \texttt{get\_error\_msg} method extracts detailed error messages from the kernel's response, ensuring that these messages are accessible for debugging purposes. The \texttt{check\_msg} method evaluates the status of the execution and prints error traces if any issues are detected, providing a clear indication of what went wrong during the execution.

\subsection{Kernel management}
The \texttt{CodeKernel} class includes several methods for managing the lifecycle of the kernel:
\begin{itemize}
    \item \texttt{shutdown}: This method stops both the backend kernel and the associated code kernel, ensuring that all resources are released.
    \item \texttt{restart}: This method restarts the kernel, providing a clean slate for subsequent code executions.
    \item \texttt{start}: This method initializes the code kernel if it is not already running, allowing for new executions.
    \item \texttt{interrupt}: This method interrupts a long-running or unresponsive kernel, providing control over runaway processes.
    \item \texttt{is\_alive}: This method checks whether the kernel is active and responsive, offering a way to monitor the kernel's status.
\end{itemize}

\section{Implementation of Knowledge Base}\label{knowledge}
The key-value knowledge base serves as a repository containing various knowledge files (e.g., Python scripts). Users can extend the knowledge base by defining their own \texttt{Knw} class, following our provided template and inheriting from the parent class. A typical \texttt{Knw} class includes the following attributes:

\begin{itemize}
	\item \texttt{name}: The name of the knowledge.
	\item \texttt{description}: A summary outlining the functionality of the knowledge, which can be function illustration, and parameters of the knowledge.
	\item \texttt{mode}: The operational mode determines how the knowledge is applied, chosen from [`core', `full'].
	\item \texttt{theta}: A threshold parameter controlling the matching difficulty between user instructions and knowledge entries (optional).
	\item \texttt{core\_function}: If \texttt{core} mode is selected, this field contains a use example illustrating how the knowledge should be applied. If \texttt{full mode} is selected, this field stores the complete implementation of the knowledge.
	\item \texttt{runnable\_function}: If the \texttt{core} mode is selected, this section contains code that can be executed directly in the backend, such as function or class definitions, among other components (optional).
\end{itemize}

Additionally, the knowledge base allows users to perform maintenance and updates, ensuring the integration remains adaptable to evolving tasks and datasets.

\begin{figure}[H]
	\centering
	\includegraphics[width=0.95\textwidth]{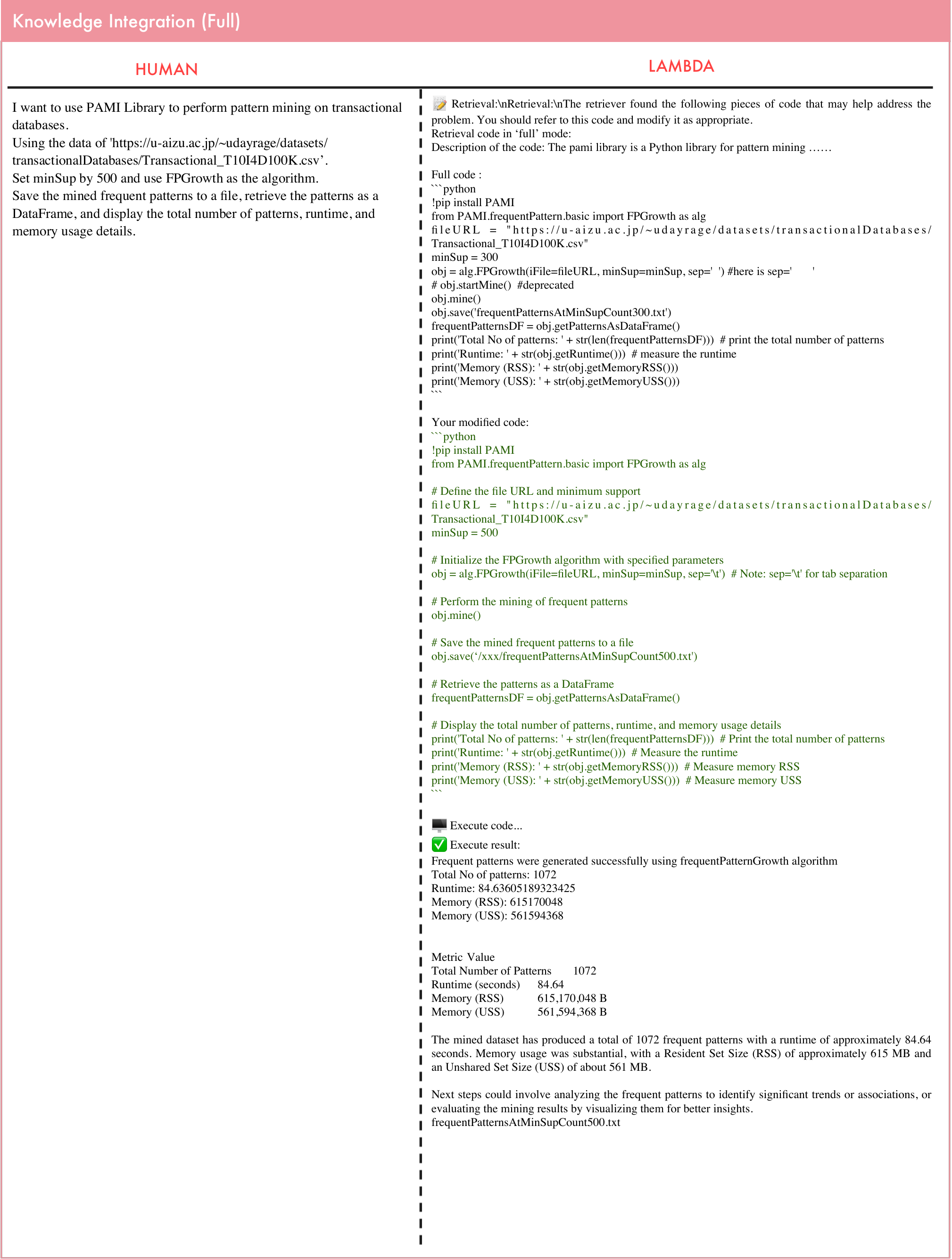}
	\caption{An illustration of integrating knowledge in 'full' mode.}\label{fig:knw_full}
\end{figure}

\begin{figure}[H]
	\centering
	\includegraphics[width=0.95\textwidth]{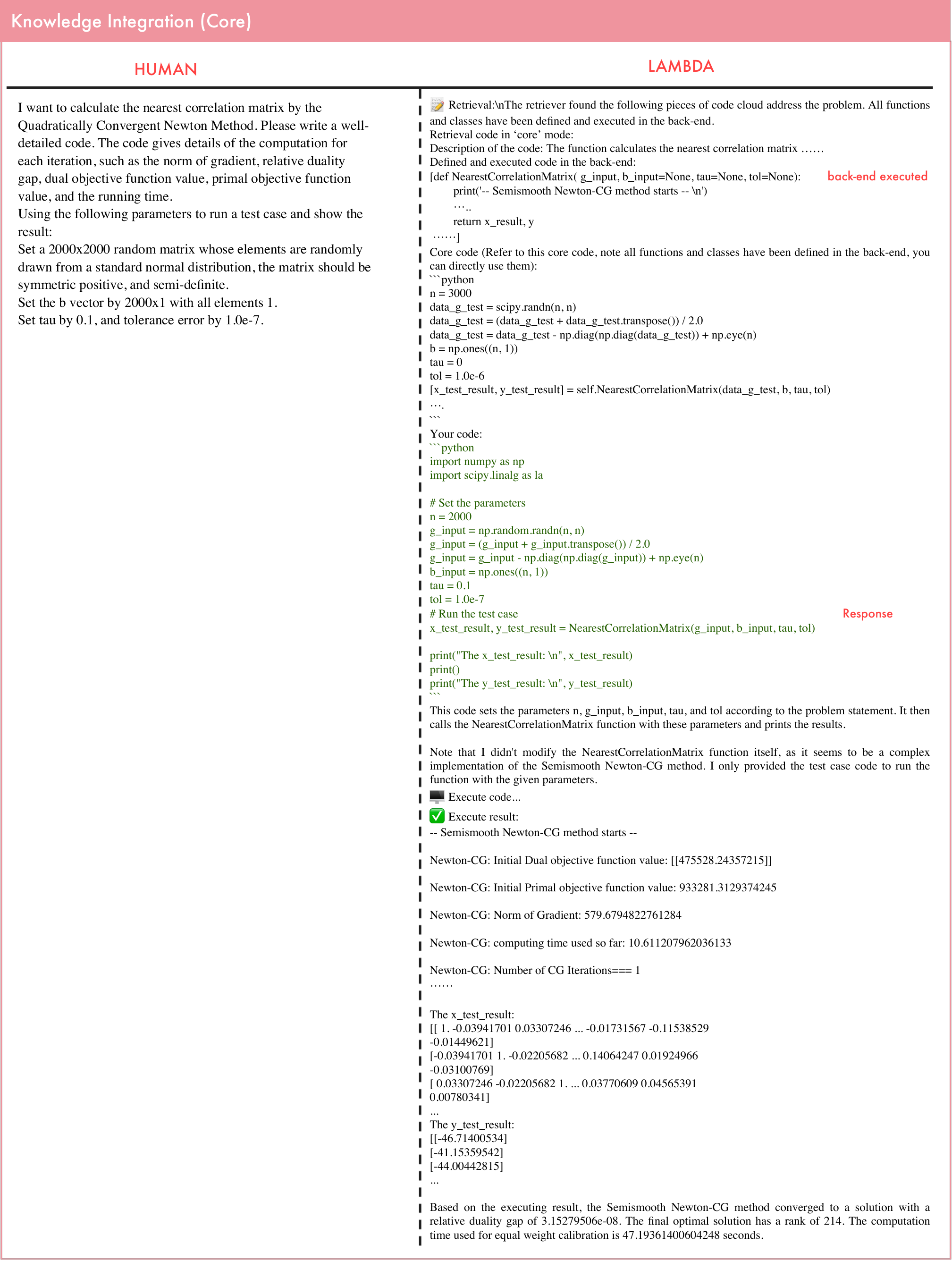}
	\caption{An illustration of integrating knowledge in 'core' mode.}\label{fig:knw_core}
\end{figure}

A more clear example of the ’full‘ and ’core‘ mode workflow, which we present two examples from our experiments, is illustrated in Figures \ref{fig:knw_full} and Figure \ref{fig:knw_core}.

\section{Discussion of LAMBDA, MetaGPT and ChatGPT-ADA}\label{disc}
The LLM-based software agents like MetaGPT are designed to automatically build software programs like calculators, games, and websites. In MetaGPT, the workflow is complex since it consists of many roles in the team like product manager, Architect, Project Manager, Engineer, and QA Engineer with much communication and collaboration in the team. So it usually costs more tokens and times to complete each task. However, the commonly used tasks in statistics and data analysis like drawing figures can be relatively easy to solve and do not need such a complicated process. To show this point, we use an example of a drawing figure. We record the time cost and token cost to show the big gap between the software agent and the data agent.

\begin{table}[h]
\centering
\begin{tabular}{@{}lrr@{}}
\toprule
         & Time   & Token  \\ \midrule
MetaGPT  & 44.75s & 19400  \\
LAMBDA   & 7.35s  & 4146   \\ \bottomrule
\end{tabular}
\caption{Time cost and token cost of a plotting task in MetaGPT and LAMBDA. The token cost is based on prompt tokens and completion tokens}
\label{tab:sa_vs_da}
\end{table}

The results from Table \ref{tab:sa_vs_da} show that software agents, such as MetaGPT, incur significant additional token usage and time when handling data analysis tasks, which is unnecessary. A more critical issue is that the output of software agents like MetaGPT consists of engineering files, such as Python scripts, which may require further manual configuration and execution by the user. In contrast, data analysis tasks demand intuitive, immediately interpretable results. Therefore, data agents offer a significant advantage in terms of interactivity within the domain of data analysis.

We also compare  LAMBDA and ChatGPT in terms of several
important features. The results are given in the Table \ref{tb:lambda_vs_gpt}.

\begin{table}[H]
    \centering
    \caption{Comparison of LAMBDA and GPT-4-Advanced Data Analysis. Scalability means the agent can integrate domain knowledge like customized models or algorithms. Portability means the agent model is adapted to other LLMs like LLaMA3 and Qwen. Human-in-the-loop means users can intervene in outcomes.}
    \label{tb:lambda_vs_gpt}
    \resizebox{0.7\textwidth}{!}{
        \begin{tabular}{lcccrrr}
            \toprule
                 & \textbf{GPT-4-Advanced Data Analysis} & \textbf{LAMBDA} \\
                \midrule
                Coding-free & \textcolor{teal}{\ding{52}} & \textcolor{teal}{\ding{52}} \\
                Scalability & \textcolor{red}{\ding{55}} & \textcolor{teal}{\ding{52}} \\
                Portability & \textcolor{red}{\ding{55}} & \textcolor{teal}{\ding{52}} \\
                Human-in-the-loop & \textcolor{red}{\ding{55}} & \textcolor{teal}{\ding{52}} \\
                Code exporting & \textcolor{red}{\ding{55}} & \textcolor{teal}{\ding{52}} \\
                Security & lower & higher \\
                Report generation & Inconvenient by prompt & Convenient \\
            \bottomrule
        \end{tabular}
    }
\end{table}

These experiments show that  LAMBDA outperforms GPT-4-Advanced Data Analysis in multiple features, highlighting the advantages of LAMBDA for data analysis tasks.

\section{Datasets}\label{data} 

Here we give the information on the sources of the datasets used in our experiments and case studies.
\begin{table}[H]
	\centering
	\caption{Datasets used in this study.}
	\label{table:datasets}
	\begin{threeparttable}
		\begin{adjustbox}{width=\textwidth,totalheight=0.4\textheight,keepaspectratio}
		\begin{tblr}{
				hline{1,17} = {-}{0.1em},
				hline{2} = {-}{},
			}
			DataSets                                                                                                                                                             & Usage                                     \\
AIDS Clinical Trials Group Study 175\tnote{1}     & Classification     \\
NHANES\tnote{2}       & Classification    \\
	Breast Cancer Wisconsin\tnote{3}  & Classification      \\
			Wine\tnote{4}    & Classification       \\
			Concrete Compressive Strength\tnote{5}      & Regression       \\
			Combined Cycle Power Plant\tnote{6}      & Regression \\
			Abalone\tnote{7}    & Regression - Education Case Study     \\
			Airfoil Self-Noise\tnote{8}   & Regression        \\
			Iris\tnote{9}       & Classification - Data Analysis Case Study \\
			Heart Disease\tnote{10}  & Regression - Education Case Study, Missing Data         \\
			Genomic Datasets \tnote{11} & High Dimensional Data \\
			Framingham Heart Study Dataset\tnote{12} & Missing Data \\
			Student Admission Records\tnote{13} & Missing Data \\
			MINIST\tnote{14} & Image Data \\
			SMS Spam\tnote{15} & Text Data \\
		\end{tblr}
		\end{adjustbox}
		\begin{tablenotes}
			\footnotesize
			\item[1]\url{https://archive.ics.uci.edu/dataset/890/aids+clinical+trials+group+study+175} \item[2]\url{https://archive.ics.uci.edu/dataset/887/national+health+and+nutrition+health+survey+2013-2014+(nhanes)+age+prediction+subset}
			\item[3]\url{https://archive.ics.uci.edu/dataset/891/cdc+diabetes+health+indicators}
			\item[4]\url{https://archive.ics.uci.edu/dataset/109/wine}
			\item[5]\url{https://archive.ics.uci.edu/dataset/165/concrete+compressive+strength}
			\item[6]\url{https://archive.ics.uci.edu/dataset/294/combined+cycle+power+plant}
			\item[7]\url{https://archive.ics.uci.edu/dataset/1/abalone}
			\item[8]\url{https://archive.ics.uci.edu/dataset/291/airfoil+self+noise}
			\item[9]\url{https://archive.ics.uci.edu/dataset/53/iris}
			\item[10]\url{https://archive.ics.uci.edu/dataset/45/heart+disease}
			\item[11]\url{https://www.kaggle.com/datasets/anhpknu/high-dimensional-data/data}
			\item[12] \url{https://www.kaggle.com/datasets/aasheesh200/framingham-heart-study-dataset/data}
			\item[13] \url{https://www.kaggle.com/datasets/mohansacharya/graduate-admissions}
			\item[14] \url{https://www.kaggle.com/datasets/hojjatk/mnist-dataset}
			\item[15] \url{https://www.kaggle.com/datasets/uciml/sms-spam-collection-dataset/data}
		\end{tablenotes}
	\end{threeparttable}
\end{table}

The Genomic datasets include the following three datasets: TCGAmirna  \citep{bentink2012}, EMTAB386  \citep{colaprico2015}, and GSE49997 \citep{pils2012validating}. A brief description of these datasets are as follows.

\begin{itemize}
\item
        The TCGAmirna dataset contains 554 patients with high-grade serous ovarian cancer, which
consists of 554 samples and 802 gene features \citep{bentink2012}.
\item
The EMTAB386 dataset contains angiogenic mRNA and microRNA gene expression signature
on 129 advanced stage, high grade serous ovarian cancers, which consists of 129 samples and
10360 gene features  \citep{colaprico2015}.
\item
The GSE49997 dataset contains the expression values of 204 epithelial ovarian cancer patients,
which consists of 194 samples and 16051 gene features  \citep{pils2012validating}.
\end{itemize}

\section{Case studies}\label{case_study}
This section presents several case studies demonstrating the use of LAMBDA, including data analysis, a self-correcting mechanism, the integration of human intelligence, its application in education, and report generation.

\begin{figure}[H]
	\begin{center}
		\includegraphics[width=0.82\textwidth]{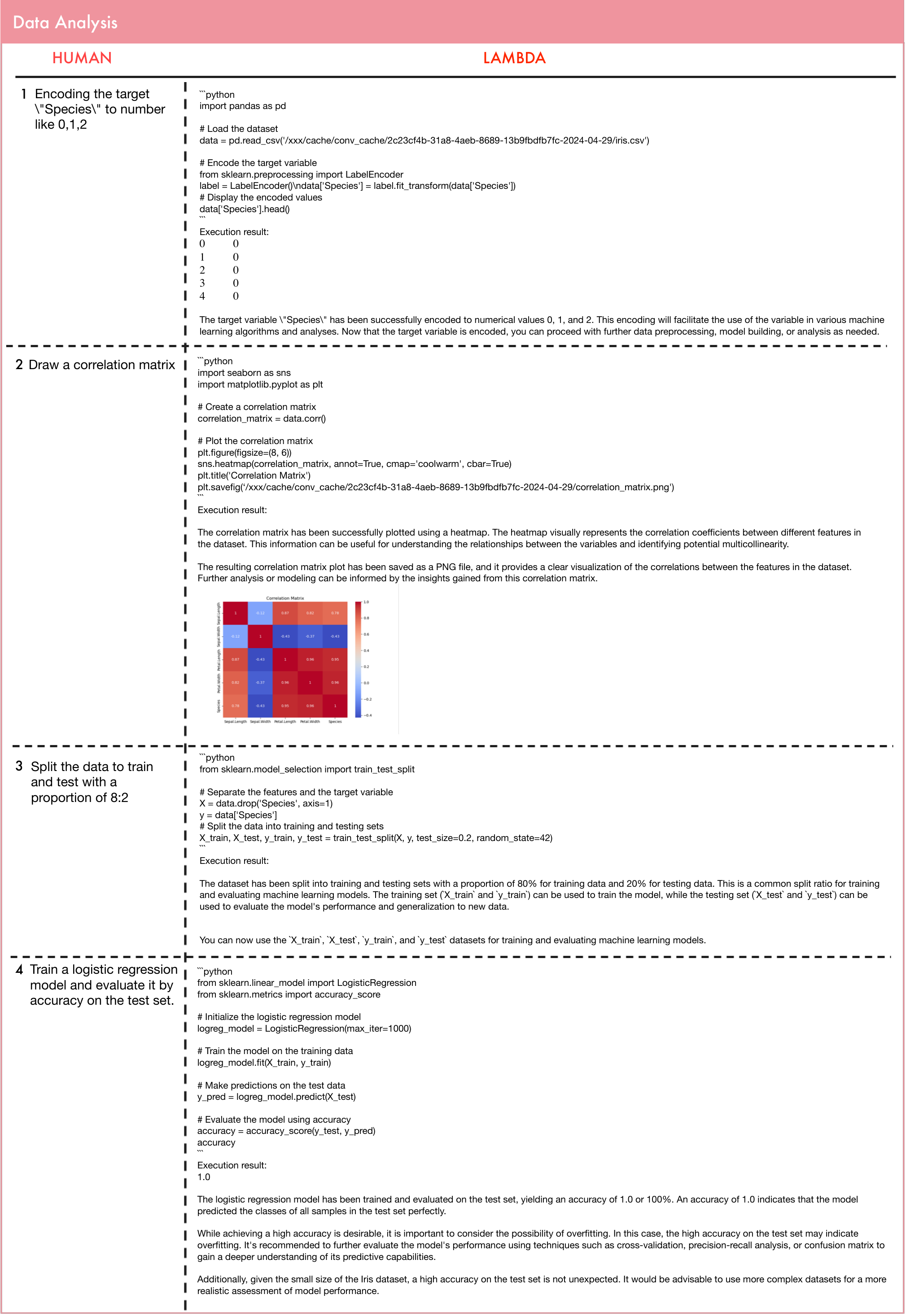}
	\end{center}
	\caption{Case study: Analysis of Iris dataset. \label{fig:cs_da1}}
\end{figure}

\begin{figure}[H]
	\begin{center}
		\includegraphics[width=0.85\textwidth]{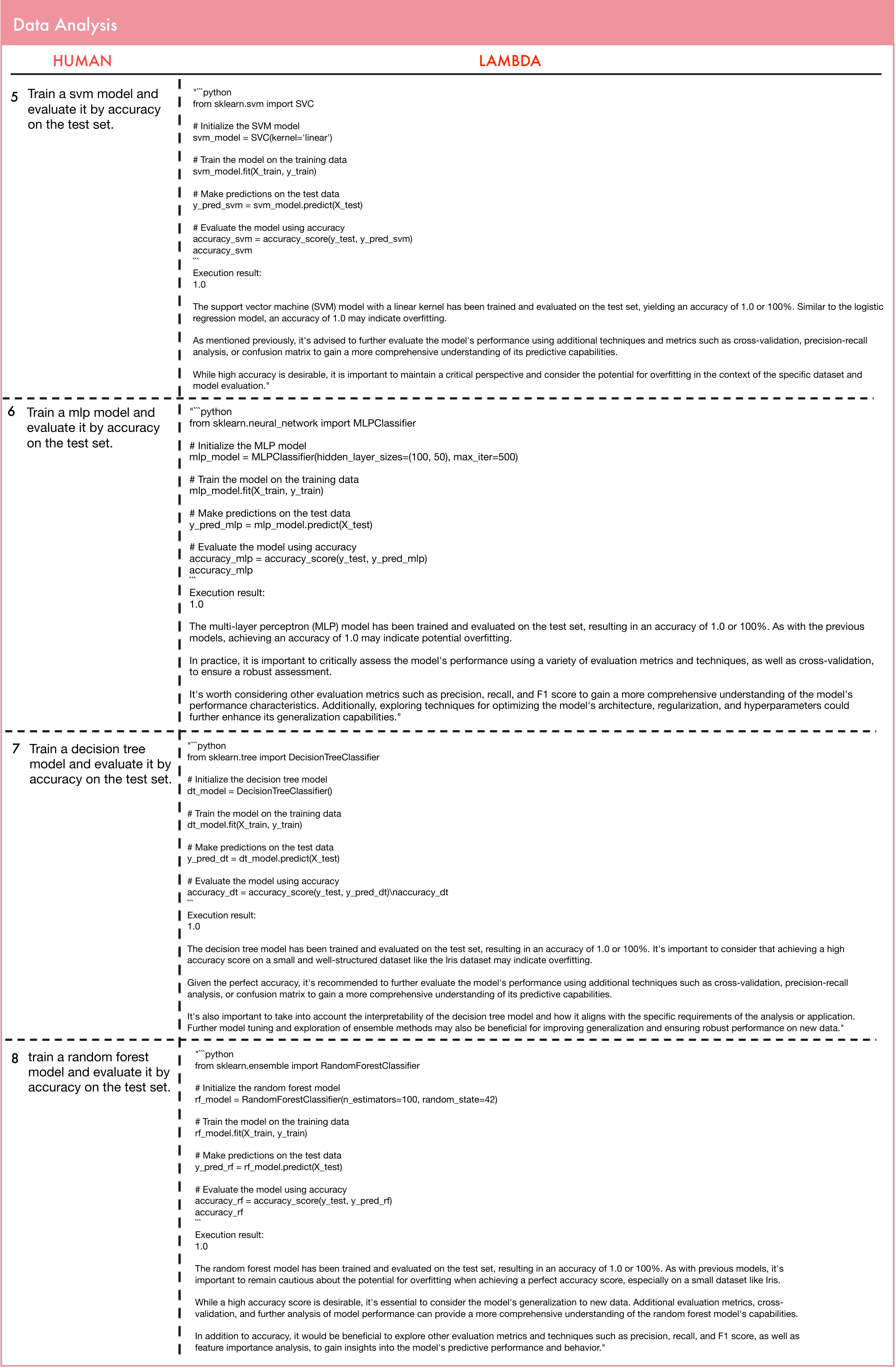}
	\end{center}
	\caption{Cont. Case study: Analysis of Iris dataset.\label{fig:cs_da2}}
\end{figure}

\begin{figure}[H]
	\begin{center}
		\includegraphics[width=\textwidth]{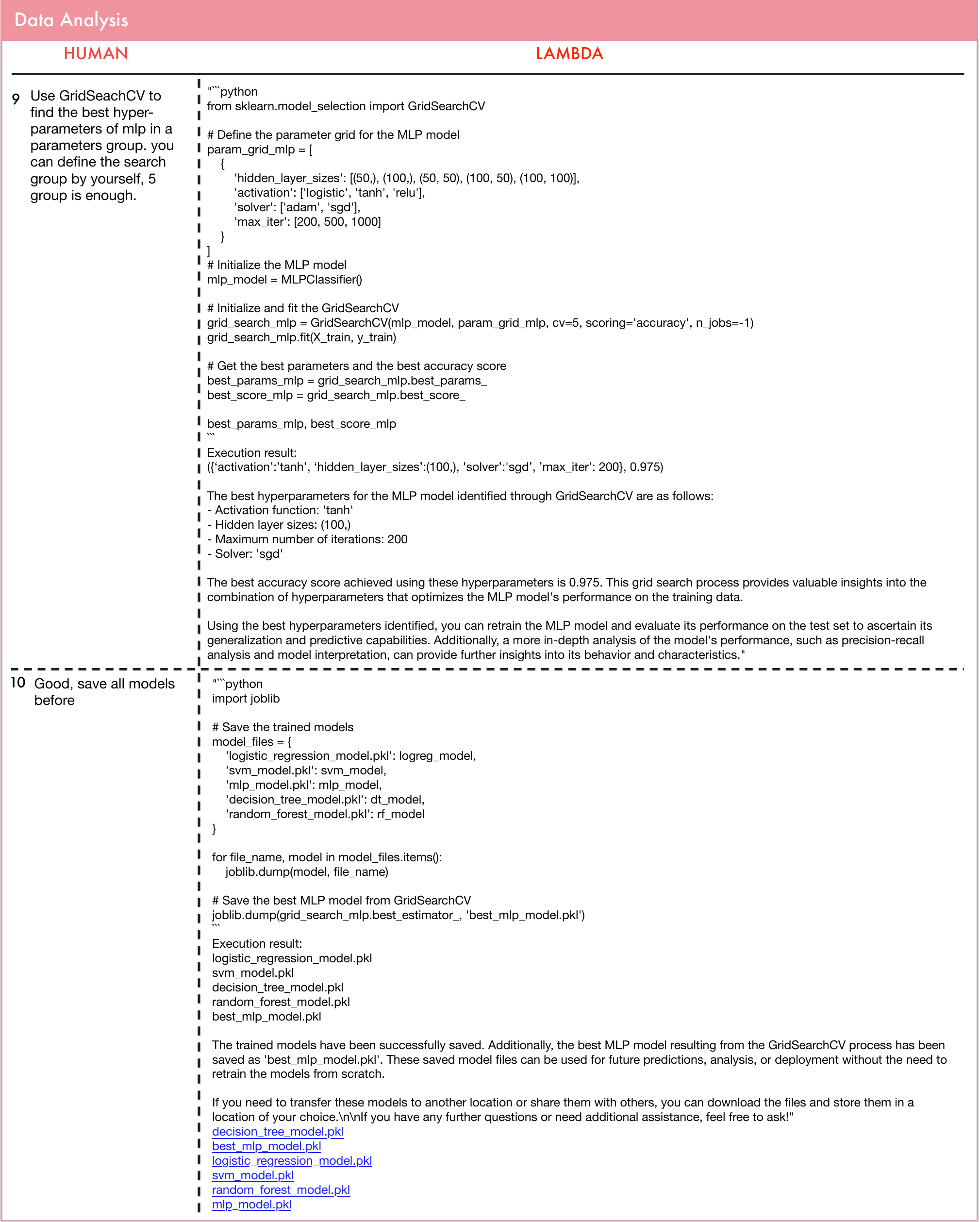}
	\end{center}
	\caption{Cont. Case study: Analysis of Iris dataset. \label{fig:cs_da3}}
\end{figure}


\begin{figure}[H]
	\begin{center}
		\includegraphics[width=\textwidth]{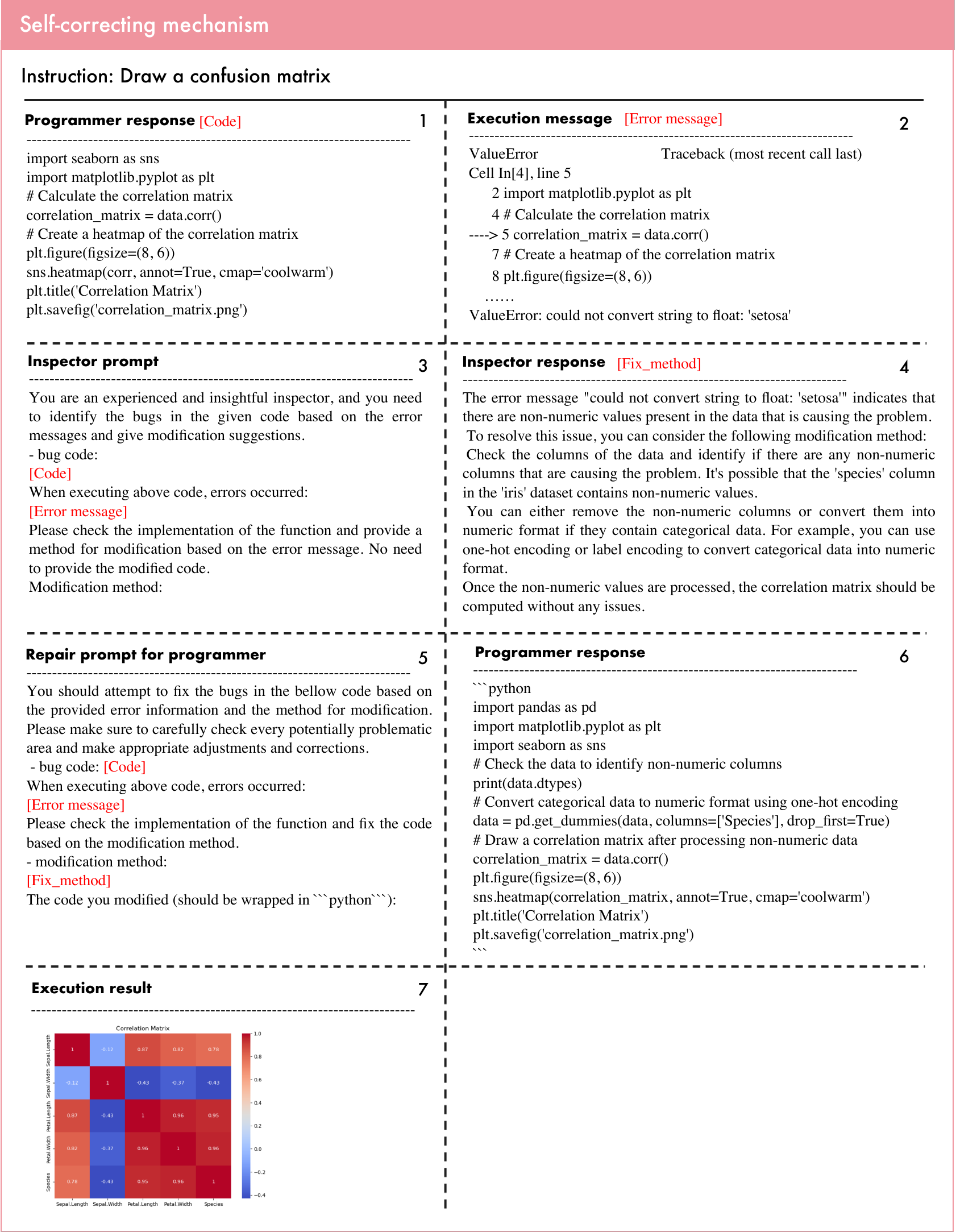}
	\end{center}
	\caption{Case study: An example of self-correcting mechanism in LAMBDA.\label{fig:self_corr}}
\end{figure}


\begin{figure}[H]
	\begin{center}
		\includegraphics[width=\textwidth]{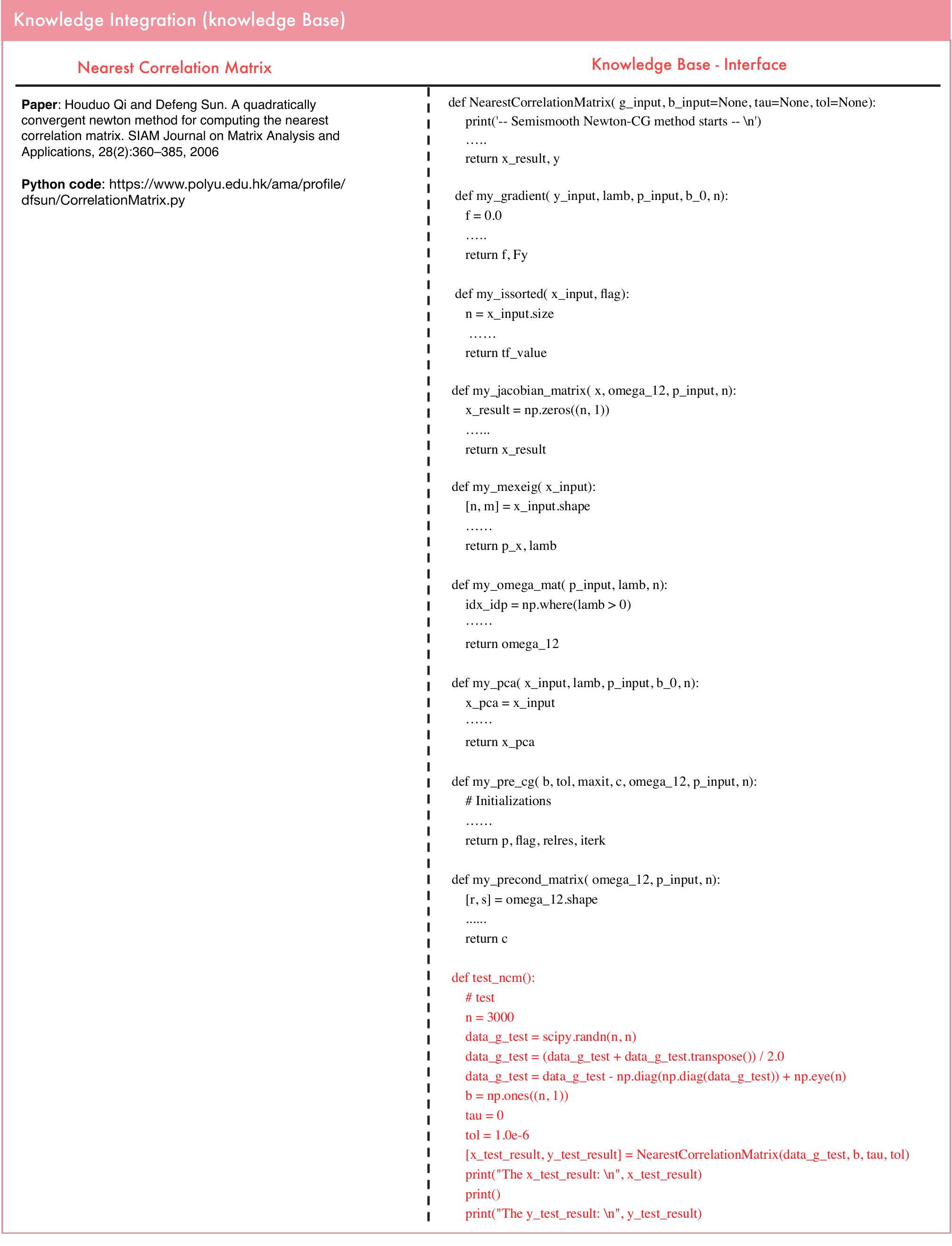}
	\end{center}
	\caption{Case study: An example of integrating human intelligence in the analysis. The red part is the `core' function in the 'Core' mode.\label{fig:cs_knw_base}}
\end{figure}

\begin{figure}[H]
	\begin{center}
		\includegraphics[width=0.95\textwidth]{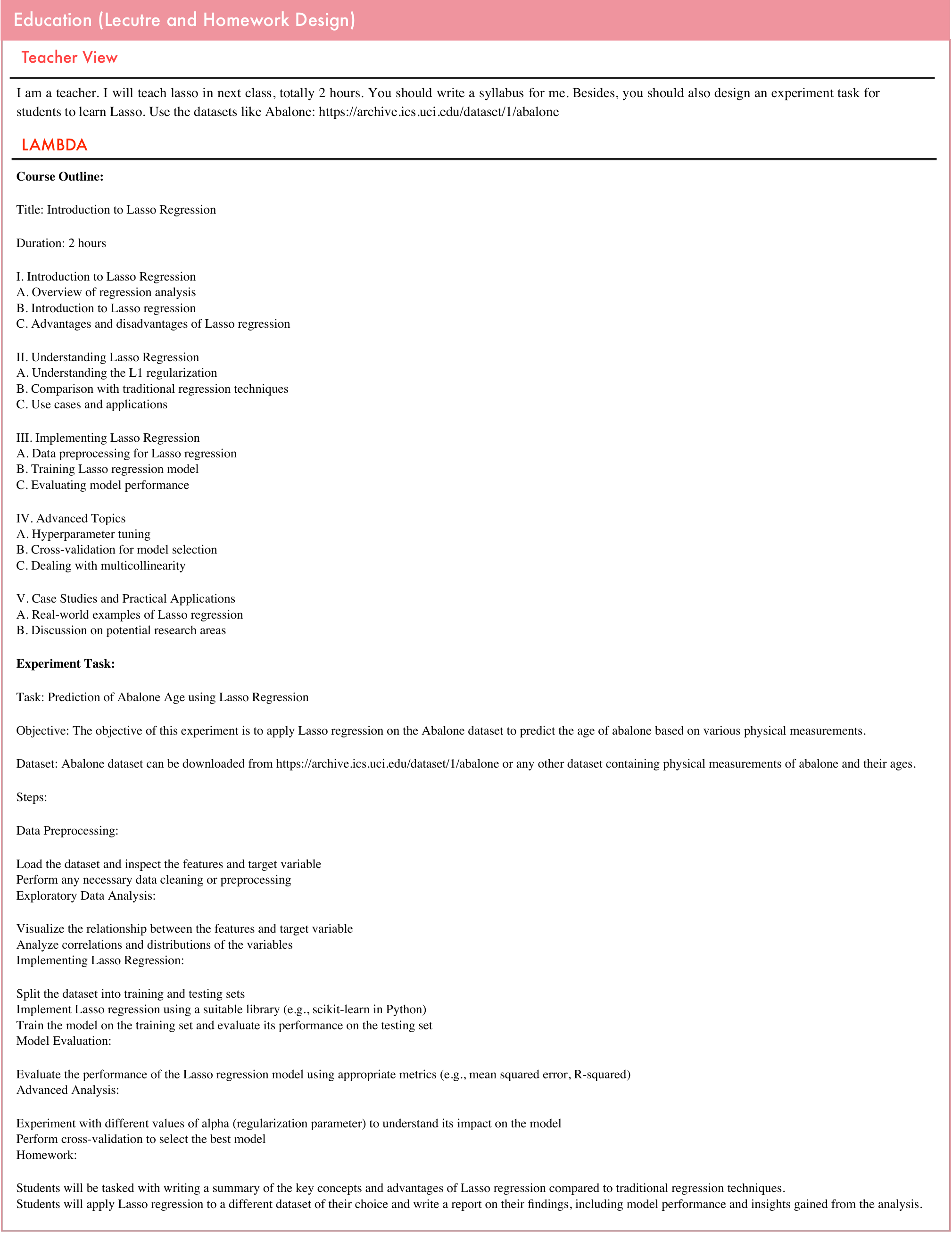}
	\end{center}
	\caption{Case study: LAMBDA in education. We use teaching Lasso as an example. \label{fig:edu_teacher}}
\end{figure}

\begin{figure}[H]
	\begin{center}
		\includegraphics[width=0.95\textwidth]{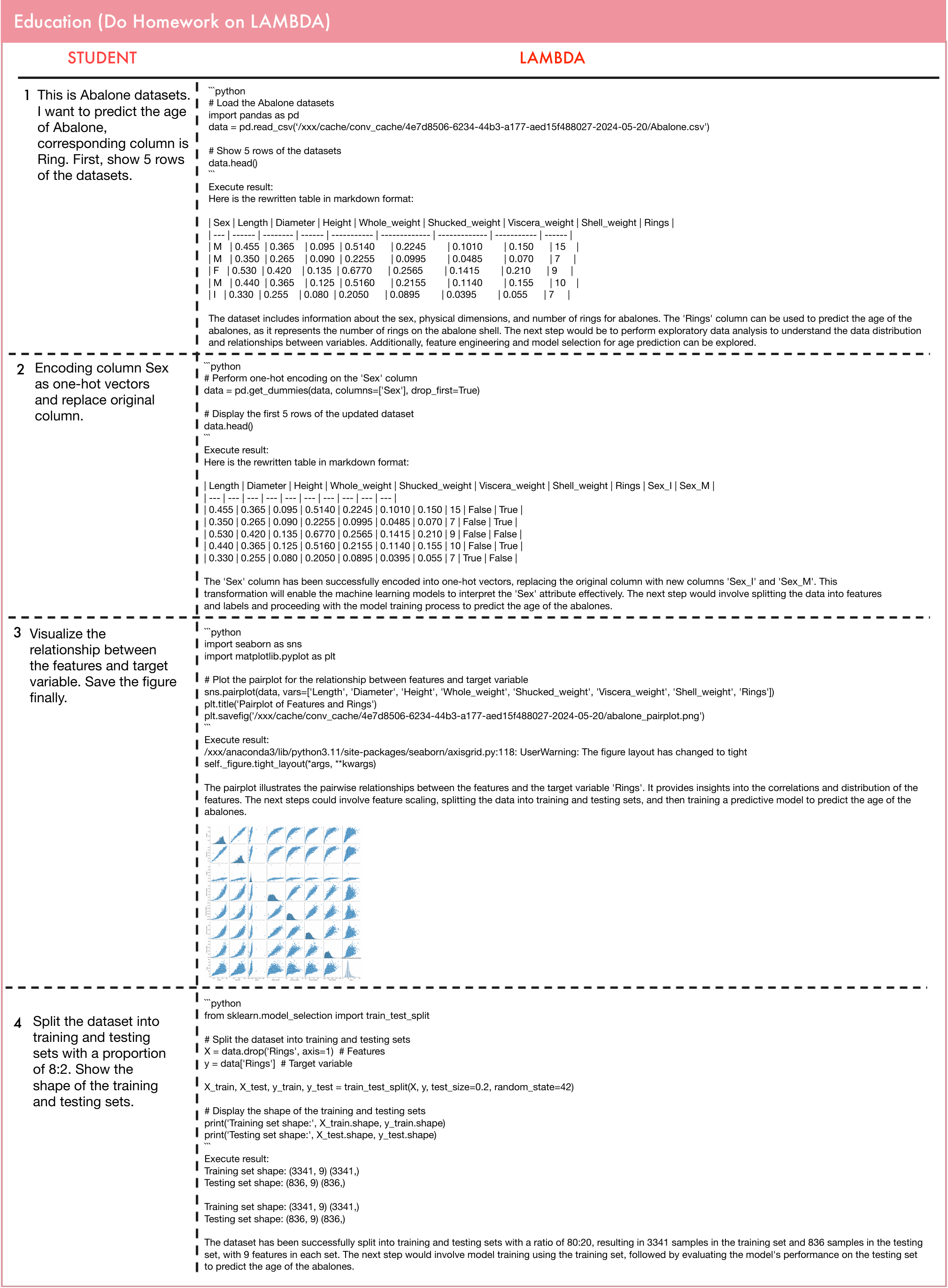}
	\end{center}
	\caption{Cont. Case Study: LAMBDA in education. Students complete assignment.
 \label{fig:cs_edu_st1}}
\end{figure}

\begin{figure}[H]
	\begin{center}
		\includegraphics[width=0.95\textwidth]{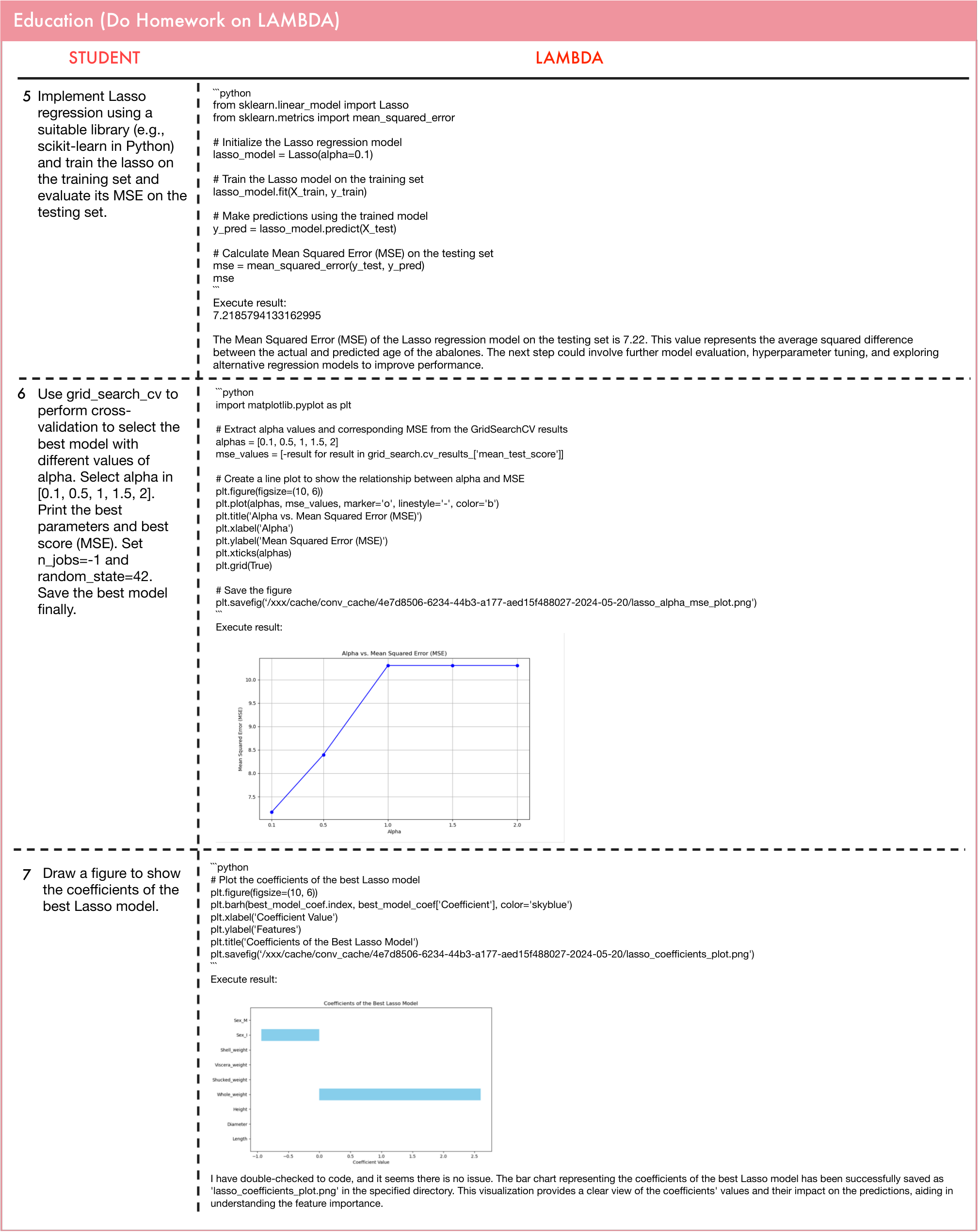}
	\end{center}
	\caption{Cont. Case study: LAMBDA in education. Students complete assignment. \label{fig:cs_edu_st2}}
\end{figure}


\begin{figure}[H]
	\begin{center}
		\includegraphics[width=\textwidth]{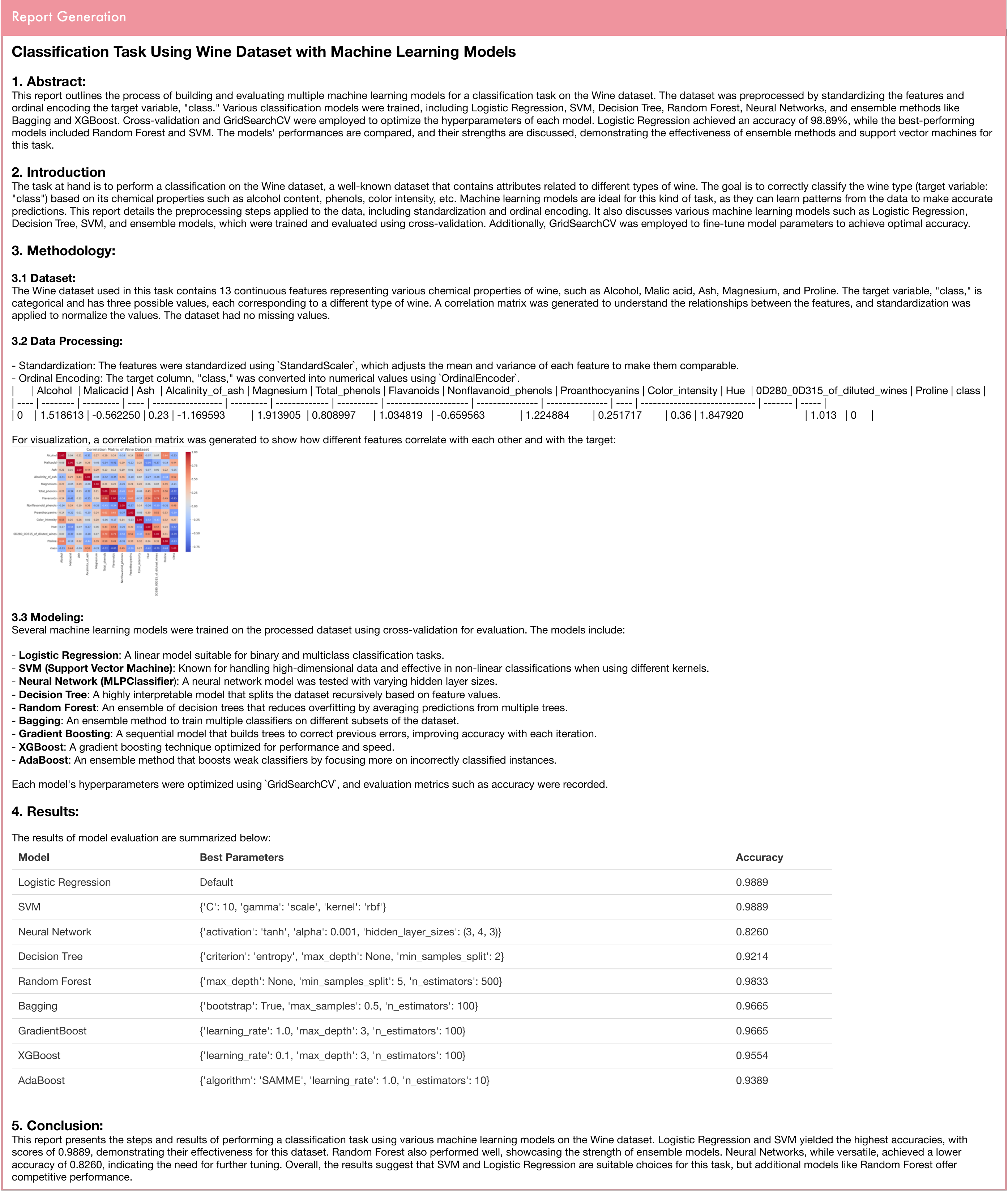}
	\end{center}
	\caption{A sample case of report generation. \label{fig:cs_report}}
\end{figure}

\section{Experimental settings}\label{exp_set}
\begin{table}[H]
	\centering
	\caption{The models and parameters used in the experiment.}
	\label{table:experiments}
	\begin{adjustbox}{width=\textwidth,totalheight=0.85\textheight,keepaspectratio}
	\rotatebox{-90}{%
		\begin{tabular}{lll}
			\hline
			\multirow{2}{*}{Experiment}                                                          & \multirow{2}{*}{Model}                             & \multirow{2}{*}{Parameters}                    \\
			&                                                    &                                                \\ \hline
			&                                                    &                                                \\
			\multirow{2}{*}{Evaluation on ML datasets}                                           & \multirow{2}{*}{Meta-Llama-3-8B-Instruct}          & \multirow{2}{*}{temperature: 0.6, top\_p: 0.9} \\
			&                                                    &                                                \\
			\multirow{2}{*}{Reliability of function Calling method}                              & \multirow{2}{*}{Qwen1.5-7B-Chat, Qwen1.5-32B-Chat} & \multirow{2}{*}{temperature: 0.7, top\_p: 0.8} \\
			&                                                    &                                                \\
			&                                                    &                                                \\
			Generating evaluation dataset for function calling                                     & Qwen1.5-110B                                       & Not given                                      \\
			Generating evaluation dataset for LAMBDA                                               & Qwen1.5-110B                                       & Not given                                      \\
			\multirow{8}{*}{Maximum number of APIs each LLM can process}                         & Meta-Llama-3-8B-Instruct                           & \multicolumn{1}{c}{\multirow{8}{*}{}}          \\
			& Llama-2-7B-chat-hf                                 & \multicolumn{1}{c}{}                           \\
			& Qwen1.5-7B-Chat                                    & \multicolumn{1}{c}{}                           \\
			& Qwen-1-8B-Chat                                     & \multicolumn{1}{c}{}                           \\
			& chatglm3-6B                                        & \multicolumn{1}{c}{}                           \\
			& chatglm2-6B                                        & \multicolumn{1}{c}{}                           \\
			& Mistral-7B-Instruct-v0.2                           & \multicolumn{1}{c}{}                           \\
			& Mistral-7B-Instruct-v0.1                           & \multicolumn{1}{c}{}                           \\
                Comparative Study of Knowledge Integration                            & GPT-3.5-turbo-1106 and specific models                              & Not given   \\			
                Case study of data analysis                                                          & GPT-3.5-turbo-1106                                 & Not given                                      \\
			Case study of integration human intelligence                                         & Meta-Llama-3-8B-Instruct                           & temperature: 0.6, top\_p: 0.9                  \\
			Case study of education                                                              & GPT-3.5-turbo-1106                                 & Not given                                      \\
			Case study of report generation                                                      & GPT-3.5-turbo-1106                                 & Not given                                      \\ \hline

		\end{tabular}%
	}
	\end{adjustbox}
\end{table}

\end{document}